\documentclass[11pt]{article}
\RequirePackage[colorlinks,citecolor=blue,urlcolor=blue,linkcolor=blue]{hyperref}
\usepackage{url}
\usepackage{booktabs}
\usepackage{svg}
\newcommand*{\email}[1]{\texttt{#1}}

\DeclareSymbolFont{slenderlargesymbols}{OMX}{ccex}{m}{n}
\DeclareMathSymbol{\prod}{\mathop}{slenderlargesymbols}{"51}
\usepackage{latexsym,amssymb,amsmath,amsthm,graphics,graphicx,float,psfrag, epsfig, color, enumerate, array}
\usepackage{amsmath}
\usepackage{amssymb}
\usepackage{mathtools}
\usepackage{amsthm}
\usepackage{authblk}

\usepackage[capitalize,noabbrev]{cleveref}

\usepackage{colortbl}
\usepackage{xcolor}
\definecolor{lightteal}{rgb}{0.7, 0.9, 0.9}

\theoremstyle{plain}
\newtheorem{theorem}{Theorem}[section]
\newtheorem{proposition}[theorem]{Proposition}

\newtheorem{corollary}[theorem]{Corollary}
\theoremstyle{definition}
\newtheorem{definition}[theorem]{Definition}

\theoremstyle{remark}

\usepackage[textsize=tiny]{todonotes}
\usepackage{latexsym,amssymb,amsmath,amsthm,graphics,graphicx,float,psfrag, epsfig, color, enumerate}

\usepackage{pgfplots}
\usepackage{url,amssymb,amsmath,amsthm,amscd,paralist,bbm,wrapfig}
\usepackage{bm}
\usepackage{textgreek}
\usepackage{algorithm}
\usepackage{algorithmic} %//format of the algorithm 

\usepackage{mathtools}

\usepackage[T1]{fontenc}
\renewcommand{\epsilon}{\varepsilon}
\newcommand{\R}{\mathbb{R}}

\newcommand{\E}{\operatorname{\mathbb{E}}}
\newcommand{\Sbb}{\operatorname{\mathbb{S}}}
\newcommand{\vol}{\operatorname{\mathrm{vol}}}

\newcommand{\Scal}{\mathcal{S}}

\newcommand{\Dcal}{\mathcal{D}}

\newcommand{\Rcal}{\mathcal{R}}
\newcommand{\Lcal}{\mathcal{L}}

\DeclareMathOperator*{\argmin}{arg\,min}
\newcommand{\prox}{\operatorname{prox}}

\begin{document}
\title{Learning Difference-of-Convex Regularizers for Inverse Problems: \\
A Flexible Framework with Theoretical Guarantees}

\author[]{Yasi Zhang and Oscar Leong}

\affil[]{Department of Statistics and Data Science, University of California, Los Angeles

            {\email{yasminzhang@ucla.edu} and \email{oleong@stat.ucla.edu}}}

\maketitle

\begin{abstract}
Learning effective regularization is crucial for solving ill-posed inverse problems, which arise in a wide range of scientific and engineering applications. While data-driven methods that parameterize regularizers using deep neural networks have demonstrated strong empirical performance, they often result in highly nonconvex formulations that lack theoretical guarantees. Recent work has shown that incorporating structured nonconvexity into neural network-based regularizers, such as weak convexity, can strike a balance between empirical performance and theoretical tractability. In this paper, we demonstrate that a broader class of nonconvex functions, difference-of-convex (DC) functions, can yield improved empirical performance while retaining strong convergence guarantees. The DC structure enables the use of well-established optimization algorithms, such as the Difference-of-Convex Algorithm (DCA) and a Proximal Subgradient Method (PSM), which extend beyond standard gradient descent. Furthermore, we provide theoretical insights into the conditions under which optimal regularizers can be expressed as DC functions. Extensive experiments on computed tomography (CT) reconstruction tasks show that our approach achieves strong performance across sparse and limited-view settings, consistently outperforming other weakly supervised learned regularizers. Our code is available at \url{https://github.com/YasminZhang/ADCR}.
\end{abstract}

\section{Introduction}

Inverse problems are ubiquitous across the natural sciences and engineering, with applications ranging from medical imaging and astronomy to compressed sensing and signal processing. In these problems, the goal is to reconstruct an unknown signal or image $x \in \R^d$ from noisy, incomplete linear measurements $y \in \R^m$: \begin{align*}
    y = Ax + \eta,
\end{align*} where $A \in \R^{m \times d}$ (with $m \leqslant d$) is a linear operator representing the measurement process and $\eta \in \R^m$ denotes noise. The inherent challenge in solving inverse problems lies in their ill-posed nature: solutions may not exist, may not be unique, or may depend discontinuously in the data $y$ \cite{Arridgeetal19}.

To address these challenges, variational regularization has emerged as a powerful and widely used framework. In this approach, one seeks a reconstruction by minimizing an objective function that balances fidelity to the observed data $y$ with a regularization term $\Rcal : \R^d \rightarrow \R$ that promotes desirable structural properties in the solution: \begin{align}
    \min_{x \in \R^d} \Lcal(x;y) + \Rcal(x).
\end{align} Here, $\Lcal(\cdot;y) : \R^d \rightarrow \R$ is a data-fidelity term that measures how well $x$ explains the observations $y$. In this work, we will mainly consider $\Lcal(x;y) := \frac{1}{2}\|Ax - y\|_2^2$, which is commonly used in many inverse problems.

Regularization strategies can be broadly characterized into two paradigms: hand-crafted and data-driven. Hand-crafted regularizers are designed based on domain knowledge to promote specific structures in the solution. Well-known examples include the $\ell_1$-norm to promote sparsity \cite{Daubechiesetal04, Tao2006, Donoho2006}, total variation \cite{Rudinetal92} to encourage piecewise smoothness, and the nuclear norm \cite{FazelThesis, Rechtetal10} to induce low-rank structure. While these regularizers are theoretically well-understood and often come with strong guarantees, they are inherently limited in flexibility. Their effectiveness relies heavily on the alignment between the assumed structure (e.g., sparsity, low-rankness) and the true underlying structure of the data. When this alignment is imperfect, hand-crafted regularizers may fail to capture the complexity of natural data. Indeed, often ``optimal'' regularizers for data distributions tend to exhibit nonconvex structure \cite{Leongetal22, alberti2021learning}.

\begin{figure*}[t]
    \centering
    \includegraphics[width=\linewidth]{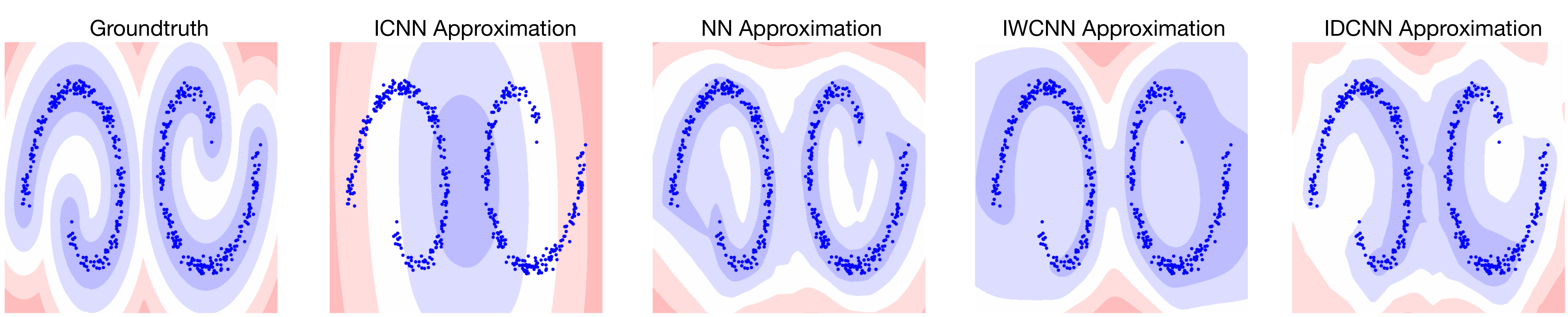}
    \caption{\textbf{Contour plots comparing the distance function to a data manifold with learned regularizers} for denoising spiral manifold data, using standard convex (ICNN), weakly convex (IWCNN), general nonconvex (NN) and difference-of-convex (IDCNN) adversarial regularization. The DC regularizer, which generalizes the weakly convex case, demonstrates improved generalization and fit to the data manifold.}
    \label{fig:first}
\end{figure*}

Data-driven regularizers, on the other hand, aim to overcome these limitations by learning a regularization functional directly from data. These methods can capture complex, data-specific structures that are difficult to encode manually. Foundational approaches in this direction include dictionary learning \cite{EladSurvey10, MairaletalSurvey14}, plug-and-play methods \cite{venkatakrishnan2013pnp, romano2017RED}, regularization using generative models \cite{Boraetal17, Handetal2018, Chungetal23}, and neural network-parametrized regularizers \cite{lunz2018adversarial, acr, shumaylov2024weakly, goujon2023neural, goujon2024learning}. Certain neural network-parametrized regularizers have other attractive properties, such as the ability to leverage unpaired data. This is in stark contrast to supervised methods, which often require paired datasets that are expensive or impractical to obtain in many applications.

Despite their success, existing data-driven regularizers often lack theoretical guarantees, particularly when the learned regularizers are nonconvex. Nonconvexity can introduce challenges in optimization and analysis, but it also offers the potential for greater flexibility and modeling power. In this work, we bridge this gap by introducing a new family of nonconvex regularization functionals that are both highly flexible and amenable to theoretical analysis.

\subsection{Outline of contributions}

We propose a framework to learn a family of flexible nonconvex regularization functionals that enjoy strong optimization guarantees. Our approach is based on parametrizing regularizers as Difference-of-Convex (DC) functions: \begin{align*}
    \Rcal_{\theta}(x) := \Rcal_{\theta_1}(x) - \Rcal_{\theta_2}(x),
\end{align*} where  $\Rcal_{\theta_1}$ and $\Rcal_{\theta_2}$ are convex. While DC functions can be highly nonconvex in general, our framework offers significant flexibility and theoretical advantages:
\begin{itemize}
    \item \textbf{Expressivity:} DC regularizers are strictly more expressive than many other nonconvex regularizers studied in the literature, such as weakly convex functions. Specifically, while every weakly convex function admits a DC decomposition, the converse is not true—DC functions can model a broader class of nonconvex structures. We demonstrate this experimentally in Figure \ref{fig:first}, where our DC regularizer achieves a better fit to ground-truth nonconvex regularization functionals compared to weakly convex alternatives.
    \item \textbf{Optimization:} The DC structure of our regularizers enables us to leverage the rich literature on DC optimization. In addition to standard gradient descent, we explore optimization algorithms that exploit the DC structure, including the Difference-of-Convex Algorithm (DCA), a classical majorization-minimization approach, and a Proximal Subgradient Method (PSM), which offer convergence guarantees under mild assumptions. 
\end{itemize}

Our specific contributions are as follows:
\begin{enumerate}
    \item We propose a flexible family of learning-based DC regularizers that are both expressive and theoretically grounded.
    \item By leveraging recent theory in star geometry and regularization, we provide conditions under which optimal regularizers in the sense of Adversarial Regularization (AR) exhibit DC structure.
    \item We propose and analyze optimization algorithms—including gradient descent, DCA, and PSM—for solving inverse problems with DC regularizers. Under appropriate assumptions, we establish convergence to limit points and stationarity guarantees.
    \item We demonstrate the practical effectiveness of our approach through extensive experiments on computed tomography (CT) reconstruction. Our method achieves strong performance in both sparse-view and limited-view settings, outperforming existing data-driven regularization techniques.
\end{enumerate}

\section{Related work}

\paragraph{Data-driven regularization:} There have been several lines of work that aim to learn regularizers in a data-driven fashion. As mentioned previously, early works include the field of dictionary learning or sparse coding \cite{MairaletalSurvey14, EladSurvey10}. Such approaches can be described as learning a polyhedral regularizer from data. Regularization by Denoising \cite{romano2017RED} constructs an explicit regularization functional using an off-the-shelf deep neural network-based denoiser. The works \cite{lunz2018adversarial, acr, shumaylov2024weakly} learn a regularizer via the critic-based adversarial loss inspired by the Wasserstein distance. The difference in these works lie in the way the neural network is parametrized to enforce certain properties, such as convexity or weak convexity. The latter \cite{shumaylov2024weakly} also establishes convergence guarantees of the Input Weakly Convex Neural Network (IWCNN) regularizer. Similarly, \cite{goujon2023neural, goujon2024learning} developed convex and weakly convex regularizers given by convolutional ridge functions with provable convergence guarantees. There have also been other works \cite{Fanetal24} that have aimed to approximate the proximal operator of a potentially nonconvex data-driven regularizer.

\paragraph{DC programming and regularization:} Our work is inspired by the rich literature on DC programming and its use in regularization. For a general survey, please see \cite{le2018dc}. In terms of DC structure in sparse regularization, several works have analyzed the use of the DC function $x \mapsto \|x\|_1 - \|x\|_2$ as a sparsity inducing regularizer \cite{ahn2017difference, yin2015minimization} as its zeros correspond to $1$-sparse vectors. Many popular nonconvex regularizers have also been shown to have a DC decomposition \cite{cao2022unifying}, such as SCAD \cite{fan2001variable}, MCP \cite{zhang2010nearly}, or the Logarithmic penalty \cite{mazumder2011sparsenet}.

\section{Preliminaries}

A function $\Rcal$ is a \textit{difference-of-convex (DC) function} if there exist convex functions $\Rcal_1$ and $\Rcal_2$ such that $$\Rcal(x) = \Rcal_1(x) - \Rcal_2(x).$$ Note that a DC decomposition is not unique as one can always consider the DC decomposition $\Rcal_1 + g - (\Rcal_2 + g)$ for any convex function $g$. This class of functions is broad and includes many interesting families of nonconvex functions, such as weakly convex functions. Indeed, recall that a function $g$ is $\rho$-weakly convex for a parameter $\rho \geqslant 0$ if the function $x\mapsto g(x) + \frac{\rho}{2}\|x\|^2_2$ is convex. Given a $\rho$-weakly convex function $g$, note that \begin{align*}
    g(x) = g(x) + \frac{\rho}{2}\|x\|_2^2 - \frac{\rho}{2}\|x\|_2^2
\end{align*} is a valid DC decomposition. DC functions are closed under many of the usual operations considered in optimization, such as by taking linear combinations of DC functions or, under certain assumptions, multiplying and dividing DC functions \cite{le2018dc}. DC functions can also be identified in a local sense, as any function that has a DC decomposition in a neighborhood of a point can be shown to be globally DC \cite{hartman59}.

We will consider optimization problems of the form \begin{align*}
    \min_{x \in \R^d} F(x) - G(x),
\end{align*} where $F,G$ are continuous and convex. We say that a function $F$ is \textit{$L$-smooth} if its gradient is $L$-Lipschitz with respect to the $\ell_2$-norm. Consider the usual \textit{subdifferential} of a convex function: $$\partial F(x) := \{g \in \R^d : F(u) \geqslant F(x) + \langle g, u - x\rangle,\ \forall u \in \R^d\}.$$ A point $x_* \in \R^d$ is a \textit{critical point} of $F - G$ if $\partial F(x_*) \cap \partial G(x_*) \neq \emptyset$, or equivalently $$0 \in \partial F(x_*) - \partial G(x_*).$$ By a slight abuse of notation, if $F(x) = \Lcal(x;y) + \Rcal_1(x)-\Rcal_2(x)$, then we will denote $\nabla F(x)$ by any vector of the form $\nabla \Lcal(x;y) + g_1 - g_2$ where $g_1 \in \partial \Rcal_1(x)$ and $g_2 \in \partial \Rcal_2(x)$.

\subsection{Learning the regularizer}
We now discuss both the architecture and learning procedure to learn the DC regularizer.

\paragraph{Input-convex neural networks:} We propose to parameterize our regularizer $\Rcal_{\theta}$ as the difference of two Input Convex Neural Networks (ICNNs) \cite{amos2017input}. ICNNs are a specialized class of neural networks designed to ensure convexity with respect to their input variables. This is achieved by enforcing non-negative weights at each layer while also using convex activation functions (e.g., ReLU or smooth variants such as Softplus). Specifically, an ICNN $\Rcal$ with $D$ layers can be described as follows: given an input $x \in \R^d$, each hidden layer output $z_i(x)$ for $i \in [D]$ is computed as \begin{align*}
    z_i(x) := \sigma_i(W_i z_{i-1}(x) + \tilde{W}_ix + b_i),
\end{align*} where $W_i$ are the weights for the previous layer's activations, $\tilde{W}_i$ are the weights for the input $x$, $b_i$ is the bias term, and $\sigma_i$ is an elementwise activation that is convex and non-decreasing. To ensure convexity, the weights $W_i$ are constrained to be non-negative, guaranteeing convexity of the map $x \mapsto \Rcal(x):=z_D(x).$ Given two ICNNs $\Rcal_1,\Rcal_2$, we call the neural network $\Rcal_1 - \Rcal_2$ an Input Difference-of-Convex Neural Network (IDCNN).

The class of ICNNs is also expressive. It has been shown \cite{chen2018optimal} that any Lipschitz convex function on a compact domain can be approximated by an ICNN parametrized via a one-layer neural network with ReLU activation and non-negative weights. One can directly apply their result to yield the following approximation guarantee for Lipschitz DC functions:

\begin{proposition}
    If $F = G - H$ is the difference of two Lipschitz convex functions over a compact domain $K$, then, for any $\epsilon > 0$, there exists an IDCNN $\Rcal_1-\Rcal_2$ such that $$\sup_{x \in K}|F(x) - (\Rcal_1(x) - \Rcal_2(x))| \leqslant \epsilon.$$
\end{proposition}
\begin{proof}
    Apply Theorem 1 in \cite{chen2018optimal} to $G$ and $H$ separately to approximate them up to an error tolerance $\epsilon/2$ with ICNNs $\Rcal_1$ and $\Rcal_2$, respectively. Using the triangle inequality yields the desired result.
\end{proof}

\paragraph{Adversarial regularization:} Using this architecture $\Rcal_{\theta} = \Rcal_{\theta_1}-\Rcal_{\theta_2}$, we propose to learn the IDCNN using the weakly supervised Adversarial Regularization (AR) framework of \cite{lunz2018adversarial}. In particular, we assume that we have access to a collection of images $x \sim \Dcal_r$ drawn from a clean data distribution modeling ground-truth images and a set of noisy measurement examples $y \sim \Dcal_y$. We also incorporate knowledge of our measurement operator by having access to a collection of ``noisy'' reconstructions with artifacts $\hat{x} \sim \Dcal_n := A^{\dagger}_{\sharp}(\Dcal_y)$, where $A^{\dagger}$ denotes the pseudo-inverse of $A$ and $f_{\sharp}(P)$ is the push-forward of a distribution $P$ under a map $f$. The fundamental philosophy of AR is that a regularizer should assign low values or high ``likelihood'' to images drawn from the clean distribution $\Dcal_r$ and assign high values or low ``likelihood'' to noisy reconstructions. This is encompassed in the following loss function: \begin{align*}
    \min_{\theta = (\theta_1,\theta_2)} \mathbb{E}_{\Dcal_r}\left[\Rcal_{\theta}(x)\right] & - \mathbb{E}_{\Dcal_n}\left[\Rcal_{\theta}(x)\right] + \lambda \cdot \mathbb{E}\left[\left(\|\nabla \Rcal_{\theta}(x) \|_2 - 1\right)_+^2\right].
\end{align*} The final term encourages the regularizer $\Rcal_{\theta}$ to be $1$-Lipschitz, and is inspired by the dual formulation of the 1-Wasserstein loss.

\section{Results on the ``Optimal'' DC Regularizer} \label{sec:star-geom}

A natural question in exploring the use of DC regularizers is whether the ideal or optimal regularizer falls under the class of DC functions. In particular, when is it the case that \begin{align*}
    \Rcal^* \in \argmin_{\Rcal \in \mathrm{Lip}(1)} \E_{\Dcal_r}[\Rcal(x)] - \E_{\Dcal_n}[\Rcal(x)]
\end{align*} is a DC function? Over the general class of Lipschitz functions, this is a challenging question to solve, but recent work \cite{Leongetal22, Leongetal24} has shown it is possible to characterize the optimal solution over the class of star body gauges (or Minkowski functionals) using tools from star geometry and Brunn-Minkowski theory \cite{geom-tom-Gardner, conv-bodies-Schneider}. In particular, we will consider regularizers of the form $x \mapsto \|x\|_K^{\alpha}$ for some power $\alpha > 0$ where \begin{align*}
    \|x\|_K := \inf\{t > 0 : x \in tK\}.
\end{align*} Here, $K$ is a \textit{star body}, i.e., a compact subset of $\R^d$ with $0 \in \mathrm{int}(K)$ such that for each $x \in \R^d \setminus \{0\}$, the ray $\{t x : t>0\}$ intersects the boundary of $K$ exactly once. Such regularizers are nonconvex for general star bodies $K$, but note that $\|\cdot\|_K$ is convex if and only if $K$ is a convex body. Any norm is the gauge of a convex body, but this class also includes nonconvex quasinorms such as the $\ell_q$-quasinorm for $q \in (0,1)$. Given such a class of functions, we are then interested in understanding when is the solution to the following problem a DC regularizer:
\begin{align*}
    \argmin_{\|\cdot\|_K^{\alpha} \in \mathrm{Lip}(1)} \E_{\Dcal_r}[\|x\|_K^{\alpha}] - \E_{\Dcal_n}[\|x\|_K^{\alpha}] .
\end{align*} 

We first give a structural result identifying the optimal regularizer for general powers of $\alpha > 0$. As discussed in the works \cite{Leongetal22, Leongetal24}, to identify the optimal regularizer, one requires a summary statistic connecting the data distribution to a corresponding data-dependent star body. For our purposes, consider a distribution $\Dcal$ with density $p$ and define the following function on the sphere: \begin{align}
    \rho_{p,\alpha}(u) := \left(\int_0^{\infty} t^{d+1-\alpha} p(tu) \mathrm{d}t\right)^{1/(d+\alpha)}. \label{eq:rho_P_alpha}
\end{align} One can show that if this function is positive and continuous over the unit sphere $\mathbb{S}^{d-1}$, it identifies a unique star body. Intuitively, this function aims to measure the average mass of a distribution and distance of this mass from the origin in any given direction on the unit sphere. Using such a quantity, we can extend the result in \cite{Leongetal24} to obtain the optimal $\alpha$-homogenous regularizer. This result is proven in the Appendix.

\begin{theorem} \label{thm:alpha-hom-theorem}
Suppose $\Dcal_r$ and $\Dcal_n$ are distributions on $\R^d$ that are absolutely continuous with respect to the Lebesgue measure with densities $p_r$ and $p_n$, respectively. Suppose $\rho_{p_r,\alpha}$ and $\rho_{p_n,\alpha}$ as defined in \eqref{eq:rho_P_alpha} are continuous over the unit sphere and that $\rho_{p_r,\alpha}(u) > \rho_{p_n,\alpha}(u) \geqslant 0$ for all $u \in \Sbb^{d-1}$. Then, there exists a star body $C_{r,n}^{\alpha}$ such that the unique minimizer of $$\E_{\Dcal_r}[\|x\|_K^{\alpha}] - \E_{\Dcal_n}[\|x\|_K^{\alpha}]$$ over all star bodies $K$ with unit volume is given by $K_{*,\alpha} := \vol_d(C_{r,n}^{\alpha})^{-1/d} C_{r,n}^{\alpha}.$ The gauge of $C_{r,n}^{\alpha}$ is given by \begin{align*}
    \|x\|_{C_{r,n}^{\alpha}} := \left(\rho_{p_r,\alpha}(x)^{d+\alpha} - \rho_{p_n,\alpha}(x)^{d+\alpha}\right)^{-1/(d+\alpha)}.
\end{align*}  
\end{theorem}

Given the precise form of the regularizer, we can then probe and ask when is this regularizer $\|\cdot\|_{K_{*,\alpha}}^{\alpha}$ a DC function. Tools from star geometry can also aid in giving concrete conditions for this. In particular, a certain type of star body addition will be useful for our purposes. Recall that for a star body $K$, its \textit{radial function} is $\rho_K(u) := \sup\{t > 0: tx \in K\}$.

\begin{definition}[\cite{Lutwak96}] \label{def:alpha-harmonic}
 For $\alpha > 0$, define the \textit{$\alpha$-harmonic radial combination between star bodies $K$ and $C$} to be the set $M_{\alpha} := M_{\alpha}(K,C)$ that satisfies \begin{align*}
        \rho_{M_{\alpha}}^{-\alpha}(u) := \rho_{K}^{-\alpha}(u) + \rho_{C}^{-\alpha}(u),\ \forall u \in \mathbb{S}^{d-1}.
    \end{align*}
\end{definition}

Due to the inverse relationship between a star body's radial function and its gauge, $\rho_K(x) = 1/\|x\|_K$ this immediately gives the following Corollary, which is our main result guaranteeing DC structure of the optimal regularizer. In certain cases, we also obtain previous results guaranteeing weak convexity of the regularizer. Please see the Appendix for the proof and examples of DC star body regularizers.

\begin{corollary} \label{cor:dc-result}
    Under the setting of Theorem \ref{thm:alpha-hom-theorem} with $\alpha \geqslant 1$, suppose $K_{*,\alpha}$ satisfies the following: there exists a convex body $C$ such that the $\alpha$-harmonic radial combination $M_{\alpha}$ between $K_{*,\alpha}$ and $C$ is a convex body. Then, the gauge $\|\cdot\|_{K_{*,\alpha}}^{\alpha}$ is a DC function. In the special case when $\alpha = 2$ and $C = \sqrt{\frac{2}{\rho}}\mathcal{B}^d$ where $\mathcal{B}^d$ is the unit Euclidean ball for some $\rho > 0$, then $\|\cdot\|_{K_{*,\alpha}}^{\alpha}$ is $\rho$-weakly convex.
\end{corollary}

\section{Algorithms for Exploiting IDCNNs} \label{sec:optimization-results}

\begin{figure*}[h]
    \centering
    \includegraphics[width=\linewidth]{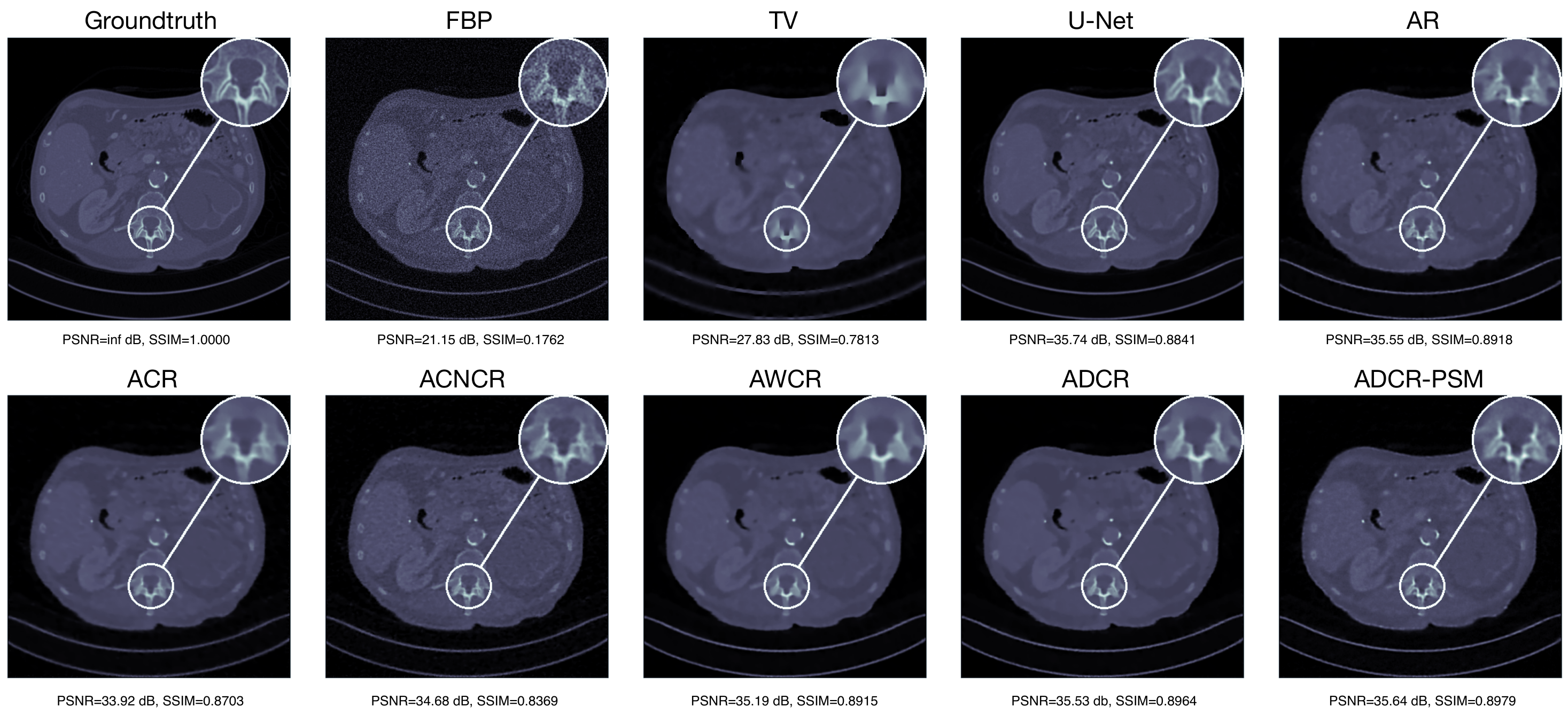}
    \caption{Qualitative comparison of reconstructed images obtained using different methods, along with the associated PSNR and SSIM, for sparse-view Computed Tomography.  ADCR and its variant, ADCR-PSM, successfully recover the fine structure of the groundtruth image  more effectively than the other weakly supervised methods, capturing intricate details that significantly enhance visual fidelity.}
    \label{fig:compare}
\end{figure*}

Given an IDCNN $\Rcal_{\theta} = \Rcal_{\theta_1} - \Rcal_{\theta_2}$, we consider solving \begin{align}
    F_* := \min_{x \in \R^d} F(x) := \Lcal(x;y) +\Rcal_{\theta_1}(x) - \Rcal_{\theta_2}(x). \label{eq:full-loss-function}
\end{align} We will focus on the case when $\Lcal(x;y) = \frac{1}{2}\|Ax-y\|_2^2$ and assume that $F_*$ is finite. We present two algorithms to exploit the variational problem's DC structure, each with strong convergence guarantees under certain assumptions.

\paragraph{Remark on assumptions:} In the subsequent results, we will exhibit guarantees when either the individual regularizers $\Rcal_{\theta_1},\Rcal_{\theta_2}$ satisfy certain smoothness conditions or the overall objective is satisfies the Kurdyka--Łojasiewicz (KL) inequality \cite{kurdyka1998gradients} (see the Appendix). The smoothness conditions can be satisfied if the ICNN uses differentiable, strictly increasing activations. Other neural network-based convex regularizers would satisfy such conditions, such as the Convex Ridge Regularizer (CRR) (see, e.g., Proposition IV.1 in \cite{goujon2023neural}). Regarding the KL assumption, this will often be satisfied. In particular, note that subanalytic functions are KL with exponent in $[0,1)$ \cite{attouch2010proximal}. The data-fit term $\Lcal(x;y) = \frac{1}{2}\|Ax-y\|_2^2$ is analytic, and for ICNNs $\Rcal_{\theta_1},\Rcal_{\theta_2}$ given by compositions of layers with piecewise, analytic activations (such as ReLU), they will be KL as well. The resulting variational objective will then be subanalytic.

\subsection{Difference-of-Convex Algorithm (DCA)}

Since the data-fit functional $\Lcal(\cdot;y)$ is convex, a natural way to solve problems of the form \eqref{eq:full-loss-function} is to linearize the concave term $-\Rcal_{\theta_2}(\cdot)$ and minimize a convex majorization of $F$: \begin{align*}
   &  x_{t+1} \in \argmin_{x \in \R^d} q(x;x_t)\ \text{where} \\
    & q(x;x_t) := \Lcal(x;y) + \Rcal_{\theta_1}(x) - \Rcal_{\theta_2}(x_t) - \langle g_t, x - x_t\rangle
\end{align*} and $g_t \in \partial \Rcal_{\theta_2}(x_t).$ This algorithm belongs to the family of majorization-minimization algorithms \cite{sun2016majorization} and is often called the Difference-of-Convex Algorithm (DCA) or Convex-Concave Procedure (CCCP) \cite{yuille2003concave}. Due to convexity of $\Lcal(\cdot;y) + \Rcal_{\theta_1}(\cdot)$, minimizing $q(\cdot;x_t)$ is a convex optimization problem. 

\begin{algorithm}
\caption{Difference-of-Convex Algorithm (DCA)}
\label{alg:dca}
\begin{algorithmic}[1]
\REQUIRE $F(x) = \Lcal(x;y) + \Rcal_{\theta_1}(x) - \Rcal_{\theta_2}(x)$
\STATE Choose an initial point $x_0 \in \mathbb{R}^d$ (e.g., $x_0 = A^{\dagger}y$)
%\ENSURE $x$ satisfying $\|x - x_*\| < \epsilon$
\FOR {$t = 0,1, 2, \ldots,T$}   \label{alg:st3}
\STATE Choose $g_t \in \partial \Rcal_{\theta_2}(x_t)$
\STATE Define $q(x;x_t): = \Lcal(x;y) + \Rcal_{\theta_1}(x) - \Rcal_{\theta_2}(x_t) -\langle g_t, x - x_t\rangle$
\STATE Compute $x_{t+1} \in \argmin_{x \in \R^d} q(x;x_t)$
\ENDFOR
\STATE \textbf{Output:} $x_T$
\end{algorithmic}
\end{algorithm}

This algorithm has attractive properties. Namely, it can be shown that when the convex part is strongly convex and differentiable and the concave part is continuously differentiable, then any limit point obtained from the sequence $(x_t)$ is stationary and $\|x_t - x_{t-1}\|_2 \rightarrow 0$ \cite{an2005dc}. We develop further guarantees here, whose main analysis is largely due to \cite{khamaru2018convergence}.

\begin{theorem} \label{thm:dca-result}
    Suppose $\Rcal_{\theta_1}$ and $\Rcal_{\theta_2}$ are continuous and convex, with $\Rcal_{\theta_1}$ $L_1$-smooth. Then Algorithm \ref{alg:dca} with iterates $(x_t)$ satisfies the following: any limit point of $(x_t)$ converges to a critical point of $F$, and the sequence of functions values $(F(x_t))$ is strictly decreasing and convergent. Moreover, \begin{align*}
            \frac{1}{T+1} \sum_{t=0}^T \|\nabla F(x_t)\|_2^2 \leqslant \frac{2(\|A\|_2^2 + L_1)(F(x_0) - F_*)}{T+1}.
        \end{align*}
\end{theorem}

\paragraph{Remark on implementation:} A challenge with DCA is that a convex optimization problem must be solved at each iteration. Empirically, however, we find that only a few iterations ($N$) of gradient descent to minimize $q(\cdot;x_t)$ at each outer iteration suffice to obtain strong performance. We perform further ablations on this inner iteration parameter $N$ for DCA and PSM in our experiments.

\subsection{Proximal Subgradient Method (PSM)}

One can also view the objective in the minimization problem \eqref{eq:full-loss-function} as the sum of a nonconvex function $\Lcal(\cdot;y) - \Rcal_{\theta_2}(\cdot)$ and a convex function $\Rcal_{\theta_1}$. Given this structure, it is also natural to consider using proximal gradient methods. For a function $\Rcal : \R^d \rightarrow \R$, recall that the proximal operator $\prox^{\Rcal}_{1/\gamma} : \R^d \rightrightarrows \R^d$ is the set-valued function \begin{align*}
    \prox^{\Rcal}_{1/\gamma}(y) := \argmin_{x \in \R^d}\  \frac{1}{2\gamma}\|y - x\|_2^2 +  \Rcal(x).
\end{align*} Then, for parameters $\alpha,\gamma > 0$, we consider the iterates \begin{align}
    x_{t+1} \in \prox_{1/\gamma}^{\Rcal_{\theta_1}}\left(x_t  - \alpha\left(\nabla \Lcal(x_t;y) - \partial \Rcal_{\theta_2}(x_t)\right)\right). \label{eq:prox-step}
\end{align}

\begin{algorithm}
\caption{Proximal subgradient method (PSM)}
\label{alg:proximal}
\begin{algorithmic}[1]
\REQUIRE $F(x) = \Lcal(x;y) + \Rcal_{\theta_1}(x) - \Rcal_{\theta_2}(x)$, $\alpha,\gamma > 0$
\STATE Choose an initial point $x_0 \in \mathbb{R}^d$ (e.g., $x_0 = A^{\dagger}y$)
%\ENSURE $x$ satisfying $\|x - x_*\| < \epsilon$
\FOR {$t = 0,1, 2, \ldots,T$}  \label{alg:st3}
\STATE Choose $g_t \in \partial \Rcal_{\theta_2}(x_t)$
\STATE Compute $x_{t+1} \in \mathrm{prox}_{1/\gamma}^{\Rcal_{\theta_1}}(x_t - \alpha (\nabla \Lcal(x_t; y) - g_t))$
\ENDFOR
\STATE \textbf{Output:} $x_T$
\end{algorithmic}
\end{algorithm}

Using tools from \cite{khamaru2018convergence}, we now turn to obtaining convergence rates for PSM. Under smoothness conditions on $\Rcal_{\theta_2}$, we can achieve similar sublinear rates to DCA, but under an additional KL assumption, this rate can be improved to be linear.

\begin{theorem} \label{thm:prox-result}
    Suppose $\Rcal_{\theta_1}$ and $\Rcal_{\theta_2}$ are continuous and convex. Then Algorithm \ref{alg:proximal} with iterates $(x_t)$, $\alpha \in (0,1/\|A\|_2^2]$, and $\gamma = 1/\alpha$ satisfies the following:
    \begin{enumerate}
        \item Any limit point of $(x_t)$ converges to a critical point of $F$, the sequence of functions values $(F(x_t))$ is strictly decreasing and convergent, and we have the bound \begin{align*}
            \frac{1}{T+1}\sum_{t=0}^T\|x_t - x_{t-1}\|_2^2 \leqslant \frac{2\alpha(F(x_0) - F_*)}{T+1} 
        \end{align*}
        \item If additionally $\Rcal_{\theta_2}$ is $L_2$-smooth, then \begin{align*}
            \frac{1}{T+1}& \sum_{t=0}^T \|\nabla F(x_t)\|_2^2 \leqslant \frac{2\alpha(\|A\|_2^2 + L_2+\alpha^{-1})^2(F(x_0) - F_*)}{T+1}. 
        \end{align*}
        \item Finally, if $F$ is KL with exponent $\omega$, then there exists a constant $c_{\omega}$ such that the linear rate holds \begin{align*}
            \frac{1}{T+1}\sum_{t=0}^T \|\nabla F(x_t)\|_2 \leqslant \frac{c_{\omega}}{T+1}.
        \end{align*}
    \end{enumerate}
\end{theorem}

\paragraph{On subgradient descent:} We note that the above result can be modified to obtain convergence guarantees for vanilla subgradient descent. In particular, one could modify the argument with the additional assumption that $\Rcal_{\theta_1}$ is $L_1$-smooth to obtain the same set of results (with $\alpha \in (0,1/(\|A\|_2^2 + L_1)]$) for subgradient descent on \eqref{eq:full-loss-function}.

\paragraph{On the proximal computation:} For convex regularizers induced by ICNNs, the proximal operator will not have a closed form. However, if one were to choose $\Rcal_{\theta_1}$ to be a hand-crafted convex regularizer with a closed-form or easy-to-compute proximal operator, then this could allow for fast algorithms to solve \eqref{eq:prox-step}. For example, the $\ell_1$-norm is a classical example of a regularizer with a closed form proximal operator, given by the soft-thresholding function. Other examples include TV, for which fast methods have been developed for TV denoising \cite{beck2009fast}. Using such regularizers would constitute a hybrid hand-crafted data-driven approach to regularizer learning and is an interesting future direction.

\begin{table}[t]
\centering
\caption{\textbf{Quantitative comparison of different methods in terms of average test PSNR and SSIM in CT experiments across both limited-view and sparse-view settings.} ADCR and its variants with different optimization algorithms achieve excellent performance across both settings in terms of PSNR and SSIM. The best scores are in \textbf{bold} and the second best are in \underline{underlined}. } 
\begin{tabular}{lcccc}
\toprule
 & \multicolumn{2}{c}{\textbf{Limited}} & \multicolumn{2}{c}{\textbf{Sparse}} \\
 \cmidrule(lr){2-3} \cmidrule(lr){4-5} 
    \textbf{Methods}             & \textbf{PSNR} & \textbf{SSIM} & \textbf{PSNR } & \textbf{SSIM} \\
 
\midrule
 
\multicolumn{2}{l}{\textit{\textbf{Knowledge-driven}}}  & & & \\
FBP & 18.5159  &  0.1781    &    21.3661 & 0.1763 \\
TV  &  25.7389 & 0.7361 & 28.5382 & 0.8048\\
%TV($\alpha=0.25$) &   26.1298 & 0.7443 & 29.7499 & 0.8332\\
% TV($\alpha=0.25, T = 500$) & 28.1936 & 0.8179 & 32.9938 & 0.8790   \\
\midrule
\multicolumn{2}{l}{\textit{\textbf{Supervised}}} & & &  \\
LPD  & 29.0123 & 0.8003 & 37.2383 & 0.9203 \\
U-Net & 29.3343 & 0.8294& 37.5111 & 0.9243\\
\midrule
\multicolumn{2}{l}{\textit{\textbf{Weakly Supervised}}}  & & & \\
AR & 26.8877 &  0.7630 &  \textbf{35.6154} & 0.9031\\
ACR &  27.1921 &  0.8101 &   33.4643 & 0.8758\\
ACNCR   &   26.9023 &  0.7932         &  33.6411 &  0.8148
\\
AWCR &   27.0888 &  0.7354 & 35.1764 & 0.8354\\
%ADCR(ours)  &    26.4418  & 0.7974 &  34.4896 & 0.8986\\
\rowcolor{lightteal} ADCR     & 27.4085 & 0.8126        &    35.5504 &  0.9060 \\
\rowcolor{lightteal} ADCR-DCA  & \underline{27.4106} & \textbf{0.8150} &  {35.5612} &   \textbf{0.9078}\\
\rowcolor{lightteal} ADCR-PSM  &  \textbf{27.4392} & \underline{0.8141} & \underline{35.5923}  & \underline{0.9075} \\

\bottomrule
\end{tabular} \label{tab:1}
\end{table}

\section{Experimental results}
We explore the effectiveness of the IDCNN through both synthetic and real experiments. We first discuss a toy example showcasing the expressivity of the IDCNN architecture in fitting to a nonconvex regularizer. Then, we demonstrate the effectiveness of our method on a computed tomography (CT) problem, comparing its performance with existing state-of-the-art weakly supervised methods.

\subsection{Distance function approximation}
To illustrate the expressivity of our architecture, we first discuss a toy example. 
We generate a synthetic double spiral dataset as shown in Figure \ref{fig:first}. The dataset comprises of 1000 points evenly divided between two interleaved spirals. Each spiral is parametrized by a random angle \(\theta\) sampled from a uniform distribution, scaled non-linearly to define the spiral radius $r$. Specifically, \( r = 2\theta + \pi \), where \(\theta \sim \sqrt{\text{Uniform}(0, 1)} \times 2\pi\). This formulation ensures a visually distinct yet intertwined structure.  
 Gaussian noise with a standard deviation $\sigma=1$  is added to the generated points. We then train an AR (i.e., a standard NN), an ACR (i.e., an ICNN), and an AWCR (i.e., an IWCNN), and ours (i.e., an IDCNN) on the denoising problem for this data, and compare the learned regularisers with the true distance function. As is clear from Figure \ref{fig:first}, the ICNN entirely fails to approximate this non-convex regularizer. This limitation is overcome by the IWCNN, and the generalization performance is further enhanced by IDCNN as theoretically the IDCNN function class contains weakly convex functions.  

\subsection{Computed Tomography (CT)}

For a fair comparison, we mainly follow the same setting as \cite{shumaylov2024weakly} and \cite{shumaylov2023provably} and all of the methods are implemented in PyTorch \cite{paszke2019pytorch} based on the official repository of CT\footnote{\url{https://github.com/Zakobian/CT_framework_/tree/master}}. The only difference between our implementation and theirs is the training/test data sampled due to randomness. We consider two applications: CT reconstruction with (i) sparse-view and (ii) limited-angle projection. In all experiments, we choose the regularization weight according to Sec 5.1 in \cite{lunz2018adversarial} without tuning. Both qualitative and quantitative results are presented in Figure \ref{fig:compare} and Table \ref{tab:1}, respectively. Our method achieves the best performance across both PSNR and SSIM \cite{ssim} evaluated, demonstrating improvements in image reconstruction over baseline methods. Implementation details are provided in the Appendix.

\begin{figure}[t]
    \centering
    \includegraphics[width=0.75\linewidth]{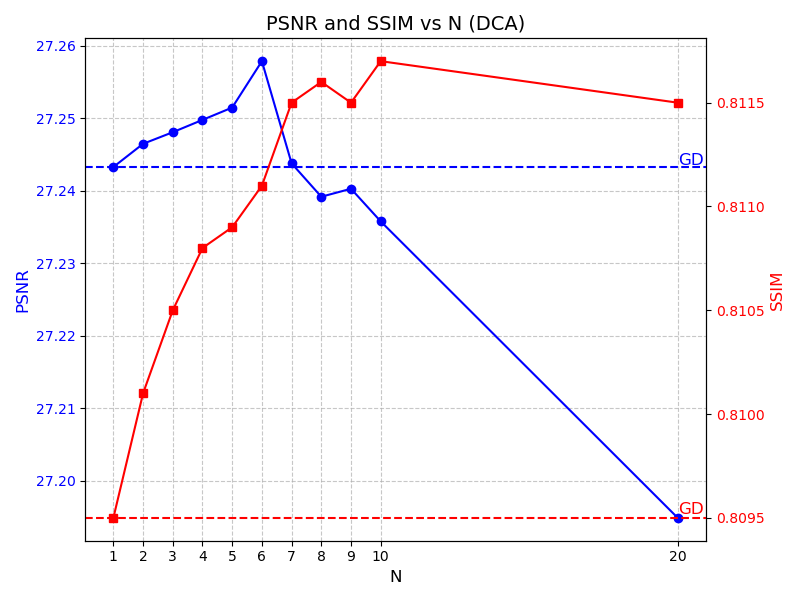}
    \caption{\textbf{Ablation on inner-loop iteration number $N$ of DCA  in the limited-view setting.} DCA  consistently improve the ADCR across various choices of the inner-loop iteration number $N$. Results in the sparse-view setting are similar. The dashed lines represent the results of ADCR obtained through gradient descent.  For DCA, we choose the optimal $N=6$ for the limited-view setting and $N=5$ for the sparse-view setting.  }
    \label{fig:abl-ccp}
\end{figure}

\paragraph{Dataset:} For the CT experiments,
human abdominal CT scans for 10 patients provided by Mayo
Clinic for the low-dose CT grand challenge \cite{moen2021low} are used.  Our training dataset
for the CT experiments consists of a total of 2250 2D slices,
each of dimension 512$\times$512, corresponding to 9 patients. The
remaining 128 slices corresponding to one patient are used
to evaluate the reconstruction performance. Projection data is simulated in ODL \cite{adler2017operator} with a GPU-accelerated astra back-end, using a parallel-beam acquisition geometry with 350 angles and 700 rays/angle, using additive Gaussian noise with $\sigma=3.2$. The pseudoinverse reconstruction is taken to be images obtained using the Filtered Back Projection (FBP).  For limited angle experiments, data is simulated with a missing angular wedge of $60^o$. 

\paragraph{Baselines:} We compare the proposed method against three types of methods: 1) knowledge-driven methods including FBP and total variation (TV) minimization \cite{Rudinetal92}, 2) supervised methods with deep neural networks including Learned Primal Dual (LPD) \cite{adler2018learned} and a standard end-to-end U-Net \cite{ronneberger2015unet}, and 3) weakly supervised methods with learned regularizers including AR \cite{lunz2018adversarial}, ACR \cite{acr}, ACNCR \cite{shumaylov2023provably}, and AWCR \cite{shumaylov2024weakly}.    

For our method, denoted by ADCR, we evaluate 3 versions: one employing the classical gradient descent to solve Eq. \eqref{eq:full-loss-function} denoted simply as {ADCR}, one using DCA, referred to as {ADCR-DCA}, and one using proximal subgradient descent, referred to as {ADCR-PSM}.  Note that supervised methods require the availability of paired clean and corrupted data, while weakly-supervised methods do not. 

\paragraph{Results:} We present both qualitative and quantitative comparisons with all the baselines in Figure \ref{fig:compare} and Table \ref{tab:1}, respectively.  Overall, ADCR achieves excellent performance across all the settings on both metrics. As demonstrated in the zooming parts of Figure \ref{fig:compare}, ADCR and ADCR-PSM successfully recover the fine structure of the groundtruth images more effectively than the other weakly supervised methods, capturing intricate details that significantly enhance visual fidelity.  Furthermore, the quantitative results in Table \ref{tab:1} corroborate these findings, showing that in the limited-view setting, ADCR consistently outperform the baseline methods in terms of PSNR and SSIM metrics. In the sparse-view setting, ADCR achieves the best SSIM score while the PSNR remains competitive, only second to AR. Notably, our proposed advanced optimization methods, DCA and PSM are both effective in improving the performance of ADCR across both settings.

\subsection{Ablation studies}
Following previous works \cite{shumaylov2023provably} and \cite{shumaylov2024weakly}, we adopt the same early stopping mechanism for the outer-loop iteration number $T$. 
We provide detailed ablation studies on the inner-loop iteration number $N$ for DCA, which is used to minimize the majorization $q(\cdot;x_t)$ in Algorithm \ref{alg:dca}, as shown in Figure \ref{fig:abl-ccp}. Additional ablations on the choice of hyperparameters in PSM are included in the Appendix.

Overall, we observe that both DCA   consistently improve the ADCR across various values of $N$.  The dashed lines represent the results of ADCR obtained through gradient descent. 
There is a clear performance improvement as $N$ initially increases. However, when $N$ continues to increase, a drop in PSNR is observed. To balance this trade-off, we select the optimal values of $N=6$ for the limited-view setting and $N=5$ for the sparse-view setting.

\section{Conclusion}
This work demonstrates the effectiveness of difference-of-convex (DC) regularization for solving ill-posed inverse problems, balancing theoretical guarantees with empirical performance. By parametrizing DC functionals as the difference of two convex components, the proposed framework enables efficient optimization using subgradient descent or DC-exploiting algorithms like DCA and PSM. We also analyze, from a star geometric perspective, the types of optimal regularizers that admit DC decompositions. Experiments on CT reconstruction tasks show superior performance in sparse and limited-view settings, outperforming baselines in retaining fine structures while maintaining strong convergence guarantees. These results highlight learning-based DC regularization as a robust and versatile tool for inverse problems.

Future work could explore advanced DC optimization methods, such as accelerated algorithms \cite{wen2018proximal}, or incorporate learning-based regularizers with efficient proximal operators into the DC framework. Additionally, it is important to mathematically characterize the nonconvex structures induced by natural data distributions and identify the families of structured nonconvex functions they belong to.

\bibliography{references}

\begin{thebibliography}{10}

\bibitem{adler2017operator}
Jonas Adler, Holger Kohr, and Ozan {\"O}ktem.
\newblock Operator discretization library (odl).
\newblock {\em Zenodo}, 2017.

\bibitem{adler2018learned}
Jonas Adler and Ozan {\"O}ktem.
\newblock Learned primal-dual reconstruction.
\newblock {\em IEEE transactions on medical imaging}, 37(6):1322--1332, 2018.

\bibitem{ahn2017difference}
Miju Ahn, Jong-Shi Pang, and Jack Xin.
\newblock Difference-of-convex learning: directional stationarity, optimality, and sparsity.
\newblock {\em SIAM Journal on Optimization}, 27(3):1637--1665, 2017.

\bibitem{alberti2021learning}
Giovanni~S Alberti, Ernesto De~Vito, Matti Lassas, Luca Ratti, and Matteo Santacesaria.
\newblock Learning the optimal tikhonov regularizer for inverse problems.
\newblock {\em Advances in Neural Information Processing Systems}, 34:25205--25216, 2021.

\bibitem{amos2017input}
Brandon Amos, Lei Xu, and J~Zico Kolter.
\newblock Input convex neural networks.
\newblock In {\em International conference on machine learning}, pages 146--155. PMLR, 2017.

\bibitem{an2005dc}
Le~Thi~Hoai An and Pham~Dinh Tao.
\newblock The dc (difference of convex functions) programming and dca revisited with dc models of real world nonconvex optimization problems.
\newblock {\em Annals of operations research}, 133:23--46, 2005.

\bibitem{Arridgeetal19}
Simon Arridge, Peter Maass, Ozan \"{O}ktem, and Carola-Bibiane Sch\"{o}nlieb.
\newblock Solving inverse problems using data-driven models.
\newblock {\em Acta Numerica}, 28:1--174, 2019.

\bibitem{attouch2010proximal}
H{\'e}dy Attouch, J{\'e}r{\^o}me Bolte, Patrick Redont, and Antoine Soubeyran.
\newblock Proximal alternating minimization and projection methods for nonconvex problems: An approach based on the kurdyka-{\l}ojasiewicz inequality.
\newblock {\em Mathematics of operations research}, 35(2):438--457, 2010.

\bibitem{beck2009fast}
Amir Beck and Marc Teboulle.
\newblock Fast gradient-based algorithms for constrained total variation image denoising and deblurring problems.
\newblock {\em IEEE transactions on image processing}, 18(11):2419--2434, 2009.

\bibitem{Boraetal17}
Ashish Bora, Ajil Jalal, Eric Price, and Alexandros Dimakis.
\newblock Compressed sensing using generative models.
\newblock {\em International Conference on Machine Learning}, 2017.

\bibitem{Tao2006}
Emmanuel~J. Cand\`{e}s, Justin~K. Romberg, and Terence Tao.
\newblock Stable signal recovery from incomplete and inaccurate measurements.
\newblock {\em Communications on Pure and Applied Mathematics}, 59(8):1207--1223, 2006.

\bibitem{cao2022unifying}
Shanshan Cao, Xiaoming Huo, and Jong-Shi Pang.
\newblock A unifying framework of high-dimensional sparse estimation with difference-of-convex (dc) regularizations.
\newblock {\em Statistical Science}, 37(3):411--424, 2022.

\bibitem{chen2018optimal}
Yize Chen, Yuanyuan Shi, and Baosen Zhang.
\newblock Optimal control via neural networks: A convex approach.
\newblock {\em arXiv preprint arXiv:1805.11835}, 2018.

\bibitem{Chungetal23}
Hyungjin Chung, Jeongsol Kim, Michael~T. Mccann, Marc~L. Klasky, and Jong~Chul Ye.
\newblock Diffusion posterior sampling for general noisy inverse problems.
\newblock {\em International Conference on Learning Representations (ICLR)}, 2023.

\bibitem{Daubechiesetal04}
Ingrid Daubechies, Michel Defrise, and Christine de~Mol.
\newblock An iterative thresholding algorithm for linear inverse problems with a sparsity constraint.
\newblock {\em Communications on Pure and Applied Mathematics}, 57(11):1413--1457, 2004.

\bibitem{Donoho2006}
David Donoho.
\newblock For most large underdetermined systems of linear equations the minimal l1-norm solution is also the sparsest solution.
\newblock {\em Communications on Pure and Applied Mathematics}, 59(6):797--829, 2006.

\bibitem{EladSurvey10}
Michael Elad.
\newblock Sparse and redundant representations: From theory to applications in signal and image processing.
\newblock {\em Springer}, 2010.

\bibitem{fan2001variable}
Jianqing Fan and Runze Li.
\newblock Variable selection via nonconcave penalized likelihood and its oracle properties.
\newblock {\em Journal of the American statistical Association}, 96(456):1348--1360, 2001.

\bibitem{Fanetal24}
Zhenghan Fan, Sam Buchanan, and Jeremias Sulam.
\newblock What's in a prior? learned proximal networks for inverse problems.
\newblock {\em International Conference on Learning Representations (ICLR)}, 2024.

\bibitem{FazelThesis}
Maryam Fazel.
\newblock Matrix rank minimization with applications.
\newblock {\em Ph.D. Thesis, Department of Electrical Engineering, Stanford University}, 2002.

\bibitem{geom-tom-Gardner}
Richard~J. Gardner.
\newblock Geometric tomography.
\newblock {\em Cambridge: Cambridge University Press}, 2006.

\bibitem{goujon2023neural}
Alexis Goujon, Sebastian Neumayer, Pakshal Bohra, Stanislas Ducotterd, and Michael Unser.
\newblock A neural-network-based convex regularizer for inverse problems.
\newblock {\em IEEE Transactions on Computational Imaging}, 2023.

\bibitem{goujon2024learning}
Alexis Goujon, Sebastian Neumayer, and Michael Unser.
\newblock Learning weakly convex regularizers for convergent image-reconstruction algorithms.
\newblock {\em SIAM Journal on Imaging Sciences}, 17(1):91--115, 2024.

\bibitem{Handetal2018}
Paul Hand, Oscar Leong, and Vlad Voroninski.
\newblock Phase retrieval under a generative prior.
\newblock {\em Advances in Neural Information Processing Systems (NeurIPS)}, 31, 2018.

\bibitem{hartman59}
Philip Hartman.
\newblock On functions representable as a difference of convex functions.
\newblock {\em Pacific Journal of Mathematics}, 9(3):707--713, 1959.

\bibitem{khamaru2018convergence}
Koulik Khamaru and Martin Wainwright.
\newblock Convergence guarantees for a class of non-convex and non-smooth optimization problems.
\newblock In {\em International Conference on Machine Learning}, pages 2601--2610. PMLR, 2018.

\bibitem{kurdyka1998gradients}
Krzysztof Kurdyka.
\newblock On gradients of functions definable in o-minimal structures.
\newblock In {\em Annales de l'institut Fourier}, volume~48, pages 769--783, 1998.

\bibitem{le2018dc}
Hoai~An Le~Thi and Tao Pham~Dinh.
\newblock Dc programming and dca: thirty years of developments.
\newblock {\em Mathematical Programming}, 169(1):5--68, 2018.

\bibitem{Leongetal24}
Oscar Leong, Eliza O'Reilly, and Yong~Sheng Soh.
\newblock The star geometry of critic-based regularizer learning.
\newblock {\em Advances in Neural Information Processing Systems (NeurIPS)}, 2024.

\bibitem{Leongetal22}
Oscar Leong, Eliza O'Reilly, Yong~Sheng Soh, and Venkat Chandrasekaran.
\newblock Optimal regularization for a data source.
\newblock {\em Foundations of Computational Mathematics}, pages 1--50, 2025.

\bibitem{lunz2018adversarial}
Sebastian Lunz, Ozan {\"O}ktem, and Carola-Bibiane Sch{\"o}nlieb.
\newblock Adversarial regularizers in inverse problems.
\newblock {\em Advances in neural information processing systems}, 31, 2018.

\bibitem{Lutwak1975}
Erwin Lutwak.
\newblock Dual mixed volumes.
\newblock {\em Pacific Journal of Mathematics}, 58(2):531--538, 1975.

\bibitem{Lutwak96}
Erwin Lutwak.
\newblock The brunn–minkowski–firey theory ii: Affine and geominimal surface areas.
\newblock {\em Advances in Mathematics}, 118(2):244 -- 294, 1996.

\bibitem{MairaletalSurvey14}
Julien Mairal, Francis Bach, and Jean Ponce.
\newblock Sparse modeling for image and vision processing.
\newblock {\em Foundations and Trends in Computer Graphics and Vision}, 8(2--3):85–283, 2014.

\bibitem{mazumder2011sparsenet}
Rahul Mazumder, Jerome~H Friedman, and Trevor Hastie.
\newblock Sparsenet: Coordinate descent with nonconvex penalties.
\newblock {\em Journal of the American Statistical Association}, 106(495):1125--1138, 2011.

\bibitem{moen2021low}
Taylor~R Moen, Baiyu Chen, David~R Holmes~III, Xinhui Duan, Zhicong Yu, Lifeng Yu, Shuai Leng, Joel~G Fletcher, and Cynthia~H McCollough.
\newblock Low-dose ct image and projection dataset.
\newblock {\em Medical physics}, 48(2):902--911, 2021.

\bibitem{acr}
S.~Mukherjee, S.~Dittmer, Z.~Shumaylov, S.~Lunz, O.~Öktem, and C.-B. Schönlieb.
\newblock Data-driven convex regularizers for inverse problems.
\newblock In {\em ICASSP 2024 - 2024 IEEE International Conference on Acoustics, Speech and Signal Processing (ICASSP)}, pages 13386--13390, 2024.

\bibitem{paszke2019pytorch}
Adam Paszke, Sam Gross, Francisco Massa, Adam Lerer, James Bradbury, Gregory Chanan, Trevor Killeen, Zeming Lin, Natalia Gimelshein, Luca Antiga, et~al.
\newblock Pytorch: An imperative style, high-performance deep learning library.
\newblock {\em Advances in neural information processing systems}, 32, 2019.

\bibitem{Rechtetal10}
Benjamin Recht, Maryam Fazel, and Pablo~A. Parrilo.
\newblock Guaranteed minimum-rank solutions of linear matrix equations via nuclear norm minimization.
\newblock {\em SIAM Review}, 52(3):471–501, 2010.

\bibitem{rockafellar2009variational}
R~Tyrrell Rockafellar and Roger J-B Wets.
\newblock {\em Variational analysis}, volume 317.
\newblock Springer Science \& Business Media, 2009.

\bibitem{romano2017RED}
Yaniv Romano, Michael Elad, and Peyman Milanfar.
\newblock The little engine that could: Regularization by denoising (red).
\newblock {\em SIAM Journal on Imaging Sciences}, 10(4):1804--1844, 2017.

\bibitem{ronneberger2015unet}
Olaf Ronneberger, Philipp Fischer, and Thomas Brox.
\newblock U-net: Convolutional networks for biomedical image segmentation.
\newblock In {\em Medical image computing and computer-assisted intervention--MICCAI 2015: 18th international conference, Munich, Germany, October 5-9, 2015, proceedings, part III 18}, pages 234--241. Springer, 2015.

\bibitem{Rudinetal92}
Leonid~I Rudin, Stanley Osher, and Emad Fatemi.
\newblock Nonlinear total variation based noise removal algorithms.
\newblock {\em Physica D: nonlinear phenomena}, 60(1-4):259--268, 1992.

\bibitem{conv-bodies-Schneider}
Rolf Schneider.
\newblock Convex bodies: The brunn–minkowski theory.
\newblock {\em Cambridge: Cambridge University Press.}, 2013.

\bibitem{shumaylov2023provably}
Zakhar Shumaylov, Jeremy Budd, Subhadip Mukherjee, and Carola-Bibiane Sch{\"o}nlieb.
\newblock Provably convergent data-driven convex-nonconvex regularization.
\newblock In {\em NeurIPS 2023 Workshop on Deep Learning and Inverse Problems}, 2023.

\bibitem{shumaylov2024weakly}
Zakhar Shumaylov, Jeremy Budd, Subhadip Mukherjee, and Carola-Bibiane Sch{\"o}nlieb.
\newblock Weakly convex regularisers for inverse problems: Convergence of critical points and primal-dual optimisation.
\newblock In {\em Forty-first International Conference on Machine Learning}, 2024.

\bibitem{sun2016majorization}
Ying Sun, Prabhu Babu, and Daniel~P Palomar.
\newblock Majorization-minimization algorithms in signal processing, communications, and machine learning.
\newblock {\em IEEE Transactions on Signal Processing}, 65(3):794--816, 2016.

\bibitem{venkatakrishnan2013pnp}
Singanallur~V Venkatakrishnan, Charles~A Bouman, and Brendt Wohlberg.
\newblock Plug-and-play priors for model based reconstruction.
\newblock In {\em 2013 IEEE Global Conference on Signal and Information Processing}, pages 945--948. IEEE, 2013.

\bibitem{ssim}
Zhou Wang, A.C. Bovik, H.R. Sheikh, and E.P. Simoncelli.
\newblock Image quality assessment: from error visibility to structural similarity.
\newblock {\em IEEE Transactions on Image Processing}, 13(4):600--612, 2004.

\bibitem{wen2018proximal}
Bo~Wen, Xiaojun Chen, and Ting~Kei Pong.
\newblock A proximal difference-of-convex algorithm with extrapolation.
\newblock {\em Computational optimization and applications}, 69:297--324, 2018.

\bibitem{yin2015minimization}
Penghang Yin, Yifei Lou, Qi~He, and Jack Xin.
\newblock Minimization of 1-2 for compressed sensing.
\newblock {\em SIAM Journal on Scientific Computing}, 37(1):A536--A563, 2015.

\bibitem{yuille2003concave}
Alan~L Yuille and Anand Rangarajan.
\newblock The concave-convex procedure.
\newblock {\em Neural computation}, 15(4):915--936, 2003.

\bibitem{zhang2010nearly}
Cun-Hui Zhang.
\newblock Nearly unbiased variable selection under minimax concave penalty.
\newblock {\em The Annals of Statistics}, 38(2):894--942, 2010.

\end{thebibliography}
\bibliographystyle{plain}

%%%%%%%%%%%%%%%%%%%%%%%%%%%%%%%%%%%%%%%%%%%%%%%%%%%%%%%%%%%%%%%%%%%%%%%%%%%%%%%
%%%%%%%%%%%%%%%%%%%%%%%%%%%%%%%%%%%%%%%%%%%%%%%%%%%%%%%%%%%%%%%%%%%%%%%%%%%%%%%
% APPENDIX
%%%%%%%%%%%%%%%%%%%%%%%%%%%%%%%%%%%%%%%%%%%%%%%%%%%%%%%%%%%%%%%%%%%%%%%%%%%%%%%
%%%%%%%%%%%%%%%%%%%%%%%%%%%%%%%%%%%%%%%%%%%%%%%%%%%%%%%%%%%%%%%%%%%%%%%%%%%%%%%
\newpage
\appendix
\onecolumn

\begin{center}
    \textbf{\Large Appendix}
\end{center}

\section{Proofs}

\subsection{Proofs for Section \ref{sec:star-geom}}

We will now prove the main Theorem on the optimal $\alpha$-homogenous regularizer. This proof will require some fundamental results in star geometry analyzing the notion of a dual mixed volume. We say that a closed set $K \subseteq \R^d$ is \textit{star-shaped} (with respect to the origin) if for all $x \in K$, we have $[0,x] \subseteq K$ where, for two points $x,y \in \R^d$, we define the line segment $[x,y] := \{(1-t)x + t y : t \in [0,1]\}.$ $K$ is a \textit{star body} if it is a compact star-shaped set such that for every $x \neq 0$, the ray $R_x := \{t x : t > 0\}$ intersects the boundary of $K$ exactly once. Equivalently, $K$ is a star body if its \textit{radial function} $\rho_K$ is positive and continuous over the unit sphere $\mathbb{S}^{d-1}$, where $\rho_K$ is defined as $
\rho_{K} (x) := \sup \{ t > 0 : t \cdot x \in K \}.$
It follows that the gauge function of $K$ satisfies $\|x\|_K = 1/ \rho_{K} (x)$ for all $x \in \mathbb{R}^{d}$ such that $x \neq 0$. We say $K$ is a \textit{convex body} if it is a compact, convex set with $0\in \mathrm{int}(K)$. We now define the notion of a dual mixed volume:
\begin{definition}[Definition 2* in \cite{Lutwak1975}]
    Given two star bodies $K,C \in \Scal^d$, the $i$-th dual mixed volume between $C$ and $K$ for $i \in \R$ is given by $$\tilde{V}_{i}(C,K) := \frac{1}{d}\int_{\Sbb^{d-1}}\rho_C(u)^{d-i} \rho_K(u)^{i}\mathrm{d}u.$$
\end{definition} One can think of dual mixed volumes as functionals that measure the size of a star body $K$ relative to another star body $C$. Note that for all $i$, $\tilde{V}_i(K, K) = \frac{1}{d}\int_{\mathbb{S}^{d-1}} \rho_K(u)^{d} \mathrm{d}u = \mathrm{vol}_d(K)$ is the usual $d$-dimensional volume of $K$. Of particular interest to us will be the case $i = -\alpha$.

\begin{proof}[Proof of Theorem \ref{thm:alpha-hom-theorem}]

 We first focus on proving the identity $$\E_{\Dcal_i}[\|x\|_K^{\alpha}] = \int_{\mathbb{S}^{d-1}} \rho_{p_i}(u)^{d+\alpha} \rho_K(u)^{-\alpha}\mathrm{d}u\ \text{for}\ i = r,n.$$  We prove this for $\Dcal_r$, since the proof is identical for $\Dcal_n$. Observe that for any $K \in \Scal^d$, since $\|u\|_K = \rho_K(u)^{-1}$, integrating in spherical coordinates gives \begin{align*}
    \E_{\Dcal_r}[\|x\|_K^{\alpha}] & = \int_{\R^d} \|x\|_K^{\alpha} p_r(x) \mathrm{d}x  = \int_{\mathbb{S}^{d-1}} \|u\|_K^{\alpha} \left(\int_0^{\infty} r^{d+\alpha-1}p_r(ru) \mathrm{d}r\right) \mathrm{d} u \\
    & = \int_{\mathbb{S}^{d-1}} \rho_K(u)^{-\alpha} \rho_{p_r,\alpha}(u)^{d+\alpha} \mathrm{d}u.
\end{align*}
    
Now, applying this identity to our objective yields \begin{align*}
        \E_{\Dcal_r}[\|x\|_K^{\alpha}] - \E_{\Dcal_n}[\|x\|_K^{\alpha}]
        & = \int_{\Sbb^{d-1}}(\rho_{p_r,\alpha}(u)^{d+\alpha} - \rho_{p_n,\alpha}(u)^{d+\alpha})\rho_K(u)^{-\alpha}\mathrm{d}u.
    \end{align*} Since $\rho_{p_r,\alpha} > \rho_{p_n,\alpha} \geqslant 0$ on the unit sphere, we have that the map $\rho_{r,n,\alpha}(u):= (\rho_{p_r,\alpha}(u)^{d+\alpha} - \rho_{p_n,\alpha}(u)^{d+\alpha})^{1/(d+\alpha)}$ is positive and continuous over the unit sphere. It is also positively homogenous of degree $-1$, as both  $\rho_{p_r}$ and $\rho_{p_n}$ are. Then, we claim that $\rho_{r,n,\alpha}$ is the radial function of the unique star body $$C_{r,n}^{\alpha}:=\{x \in \R^d : \rho_{r,n,\alpha}(x)^{-1} \leqslant 1\}.$$ We first show that $C_{r,n}^{\alpha}$ is star-shaped with gauge $\|x\|_{C_{r,n}^{\alpha}} = \rho_{r,n,\alpha}(x)^{-1}$. Note that the set is star-shaped since for each $x \in C_{r,n}^{\alpha}$, we have for any $t \in [0,1]$, $$\rho_{r,n,\alpha}(tx)^{-1} = (\rho_{r,n,\alpha}(x)/t)^{-1} = t \rho_{r,n,\alpha}(x)^{-1} \leqslant t \leqslant 1$$ since $\rho_{r,n,\alpha}$ is positively homogenous of degree $-1$. Hence $[0,x] \subseteq C_{r,n}^{\alpha}$ for any $x \in C_{r,n}^{\alpha}$. Then, we have that the gauge of $C_{r,n}^{\alpha}$ satisfies \begin{align*}
     \|x\|_{C_{r,n}^{\alpha}} & = \inf\{t : x/t \in  C_{r,n}^{\alpha}\} = \inf\{t : \rho_{r,n,\alpha}(x/t)^{-1} \leqslant 1\} \\
     & = \inf\{t : \rho_{r,n,\alpha}(x)^{-1} \leqslant t\} = \rho_{r,n,\alpha}(x)^{-1}.
 \end{align*} Hence the gauge of $C_{r,n}^{\alpha}$ is precisely $\rho_{r,n,\alpha}(x)^{-1}$. Moreover, by assumption, $u \mapsto \rho_{r,n,\alpha}(u)$ is positive and continuous over the unit sphere so $C_{r,n}^{\alpha}$ is a star body and is uniquely defined by $\rho_{r,n,\alpha}$.
 
Combining the above results, we have the dual mixed volume interpretation \begin{align*}
        \E_{\Dcal_r}[\|x\|_K^{\alpha}] - \E_{\Dcal_n}[\|x\|_K^{\alpha}] &  = \int_{\Sbb^{d-1}}\rho_{r,n,\alpha}(u)^{d+\alpha}\rho_K(u)^{-\alpha}\mathrm{d}u = d \tilde{V}_{-\alpha}(C_{r,n}^{\alpha}, K).
    \end{align*} Theorem 2 in \cite{Lutwak1975} states that for any two star bodies $K,C \in \Scal^d$ and any $i,j,k \in \R$ such that $i < j < k$, $$\tilde{V}_{j}^{k-i}(K,C) \leqslant \tilde{V}_i^{k-j}(K,C) \tilde{V}_{k}^{j-i}(K,C),$$ with equality if and only if $K$ and $C$ are dilates, i.e., $K = \lambda C$ for some $\lambda > 0$. Setting $i = -\alpha,  j = 0,$ and $k = d$ gives $$\E_{\Dcal_r}[\|x\|_K^{\alpha}] - \E_{\Dcal_n}[\|x\|_K^{\alpha}] = d\tilde{V}_{-\alpha}(C_{r,n}^{\alpha},K) \geqslant d\vol_d(C_{r,n}^{\alpha})^{(d+\alpha)/d}\vol_d(K)^{-\alpha/d}$$ with equality if and only if $K$ is a dilate of $C_{r,n}^{\alpha}.$ Thus, the objective is minimized over the collection of unit-volume star bodies by $K_{*,\alpha} := \vol_d(C_{r,n}^{\alpha})^{-1/d}C_{r,n}^{\alpha}.$
    
\end{proof}

\begin{proof}[Proof of Corollary \ref{cor:dc-result}]
We first highlight the following fact concerning the radial function and gauge: given a star body $K$, we have that $\rho_K(x) = 1/\|x\|_K$. Hence we can relate the $\alpha$-harmonic radial combination $M_{\alpha}(K,C)$ between star bodies $K$ and $C$ to their gauge functions for all $u \in \mathbb{S}^{d-1}$:
\begin{align*}
    \rho_{M_{\alpha}(K,C)}^{-\alpha}(u) = \rho_{K}^{-\alpha}(u) + \rho_{C}^{-\alpha}(u) \Longleftrightarrow \|u\|_{M_{\alpha}(K,C)}^{\alpha} = \|u\|_K^{\alpha} + \|u\|_C^{\alpha}.
\end{align*}

Under the assumptions of the Corollary, we have that there exists a convex body $C$ such that the $\alpha$-harmonic radial combination $M_{\alpha}$ of $K_{*,\alpha}$ and $C$ is convex. Recall that for a non-negative convex function $f$ and a non-decreasing convex function $\phi$, the composition $\phi(f(\cdot))$ is convex. Since $M_{\alpha}$ is convex and $\alpha \geqslant 1$, this means that the gauge $\|\cdot\|_{M_{\alpha}}^{\alpha}$ is a convex function.  By the same logic, $\|\cdot\|_C^{\alpha}$ is convex. But by definition of the $\alpha$-harmonic radial combination, this means that $\|\cdot\|_{K_{*,\alpha}}^{\alpha}$ admits the DC decomposition $$ \|x\|_{K_{*,\alpha}}^{\alpha} = \|x\|_{M_{\alpha}}^{\alpha} - \|x\|_C^{\alpha}.$$ This concludes the proof for the first half of the Corollary. 

For the second half, recall that for $\lambda > 0$ and any star body $K$, $\|x\|_{\lambda K} = \lambda^{-1}\|x\|_K$. When $\alpha = 2$, note that if $C = \sqrt{\frac{2}{\rho}} \mathcal{B}^d$, then $\|x\|_C^2 = \|x\|^2_{\sqrt{\frac{2}{\rho}} \mathcal{B}^d} = \frac{\rho}{2}\|x\|_2^2.$ By our previous argument, this means that $$\|x\|_{K_{*,2}}^2 = \|x\|_{M_2}^2 - \frac{\rho}{2}\|x\|_2^2.$$ By convexity of the function $\|\cdot\|_{M_2}^2$, this implies that $\|\cdot\|_{K_{*,2}}^2$ is $\rho$-weakly convex since \begin{align*}
   x \mapsto \|x\|_{K_{*,2}}^2 + \frac{\rho}{2}\|x\|_2^2 = \|x\|_{M_2}^2 - \frac{\rho}{2}\|x\|_2^2 - \frac{\rho}{2}\|x\|_2^2 = \|x\|_{M_2}^2
\end{align*} is convex.
\end{proof}
\paragraph{Examples of DC star regularizers:} As an example from the literature, note that the perturbed $\ell_{1-2}$ regularizer \cite{yin2015minimization} $\|\cdot\|_1 - \rho\|\cdot\|_2$ for $\rho \in (0,1)$ is a valid DC star body regularizer. The condition $\rho < 1$ is simply necessary to ensure the resulting star body has non-empty interior. However, note that Definition \ref{def:alpha-harmonic} gives us an explicit way to build DC star body regularizers. In particular, one can always take two convex bodies $C$ and $M$ and check whether $\|u\|_M > \|u\|_C$ for all $u \in \mathbb{S}^{d-1}$ by positive homogeneity. In terms of sets, this means that $M$ is \textit{strictly contained} in $C$, in the sense that $M \subset \mathrm{int}(C).$

\begin{figure*}[h]
    \centering
    \includegraphics[width=0.75\linewidth]{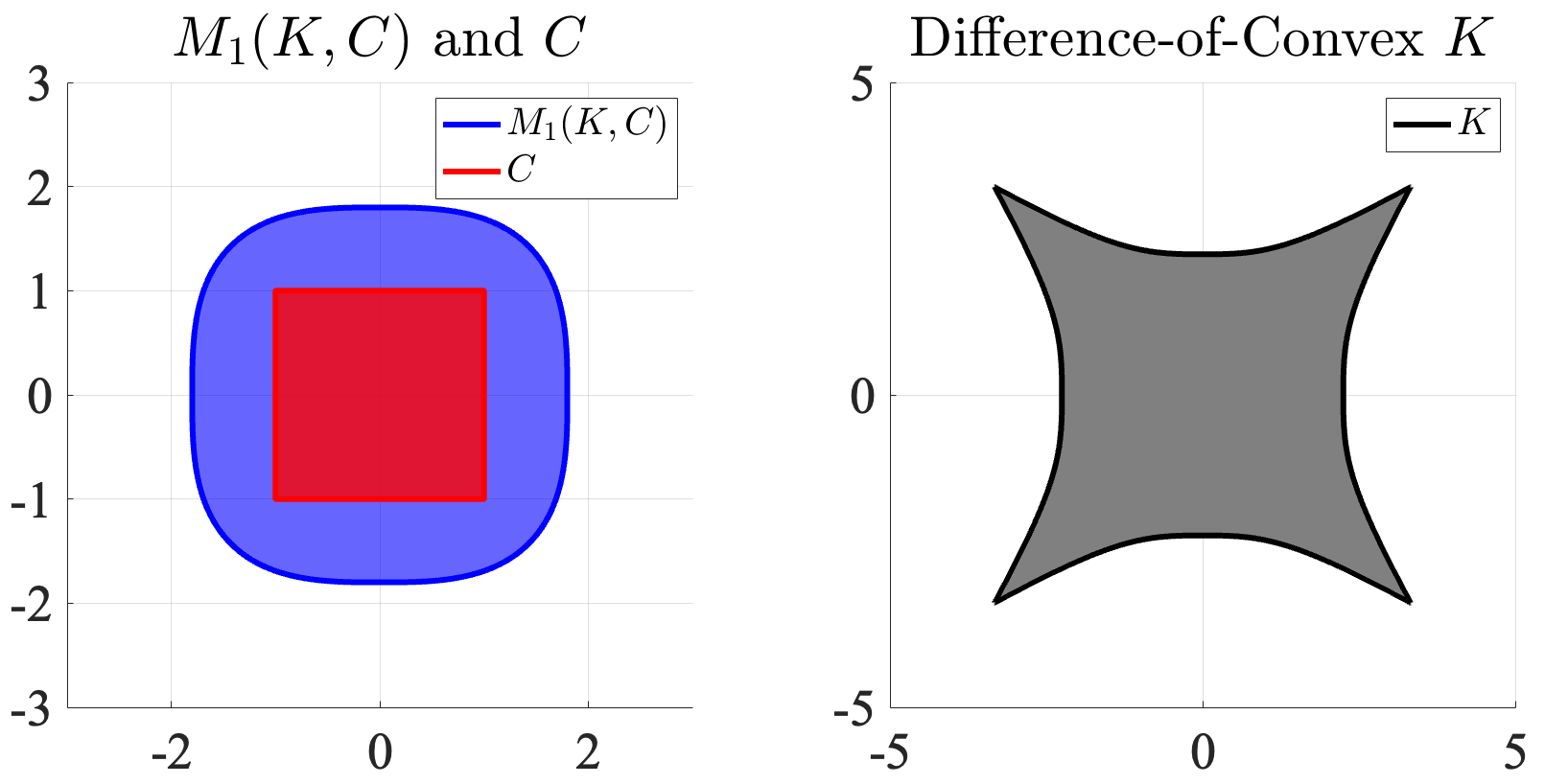}
    \caption{We visualize the geometry of a DC regularizer's level set in terms of its individual convex components. In particular, the gauge of $K$ can be written as the difference of two convex gauges, induced by the sets $M_1$ and $C$. $M_1$ can also be understood as a particular type of star body addition between $K$ and $C$. In directions where the boundaries $M_1$ and $C$ are close, $K$ is ``spiky'' and its boundary is further from the origin.}
    \label{fig:dc-star-reg}
\end{figure*}

As a visual example in $d=2$, consider the $\ell_{\infty}$-ball $M_1 := \{x \in \R^d : \|x\|_{\infty} = \max_{i \in [d]}|x_i| \leqslant 1\}$ and a scaled $\ell_3$-norm ball $C := \{x \in \R^d : \|x\|_3 = \left(\sum_{i=1}^d|x_i|^3\right)^{1/3} \leqslant 1.8\}.$ Then define $K := \{x \in \R^d : \|x\|_{M_1} - \|x\|_{C} \leqslant 1\}.$ Note the suggestive notation: by construction, $M_1(K,C)$ is the $1$-harmonic radial combination between $K$ and $C$. We depict the shapes of these sets in Figure \ref{fig:dc-star-reg}. Based on this figure, we can see how the geometry of $K$ relates to $M_1$ and $C$. In particular, note that for directions in which the boundaries of $M_1$ and $C$ are close, the boundary of $K$ is far from the origin and exhibits spikier behavior. Likewise, in directions where the boundaries are further apart, $K$ is also closer to the origin. From a regularization perspective, the directions in which $K$ is ``spiky'' would be those that the regularizer prefers as the gauge is small in such directions. For example, the $\ell_{1-2}$ regularizer prefers sparse vectors as its zeros exactly coincide with $1$-sparse vectors, i.e., $\|x\|_1 - \|x\|_2 = 0 \Longleftrightarrow \|x\|_0 = 1$. This useful ``spiky'' geometry is also akin to the convex geometric perspective of sparse vectors lying on low-dimensional faces of the $\ell_1$-ball.

\subsection{Proofs for Section \ref{sec:optimization-results}}
We now provide proofs for our optimization guarantees. The proof techniques are based on previous results in the literature (e.g., \cite{khamaru2018convergence}) and we provide proofs here for completeness. We first require certain definitions: 

\begin{definition}
    Consider a function $F : \R^d \rightarrow \R$. We say that $F$ satisfies the \textit{Kurdyka--Łojasiewicz (KL) inequality} at $\overline{x} \in \R^d$ if there exists a scalar $\omega \in [0,1)$ and a constant $c > 0$ such that the following inequality holds for all $x$ a neighborhood of $\overline{x}$: \begin{align*}
        \|\nabla F(x)\|_2 \geqslant c|F(x) - F(\overline{x})|^{\omega}.
    \end{align*} If $F$ satisfies this inequality for all $\overline{x}$ with the same exponent $\omega$, then we say that $F$ satisfies the \textit{KL inequality with exponent} $\omega$ or is a \textit{KL function with exponent} $\omega$.
\end{definition}
\begin{proof}[Proof of Theorem \ref{thm:dca-result}]
    For notational convenience, let $\tilde{\Lcal}(x):= \Lcal(x;y) + \Rcal_{\theta_1}(x)$ and set $q(x;x_t) := \tilde{\Lcal}(x) - \Rcal_{\theta_2}(x_t) - \langle g_t, x - x_t\rangle$ where $g_t \in \partial \Rcal_{\theta_2}(x_t)$. We will first establish two preliminary facts:
    \begin{enumerate}
        \item \textbf{Smoothness of $\tilde{\Lcal}$:} We have that $\tilde{\Lcal}$ is $\tilde{L}:=(\|A\|_2^2 + L_1)$-smooth since for any $x,z \in \R^d$, using the triangle inequality and submultiplicativity of the norm, we have \begin{align*}
        \|\nabla \tilde{\Lcal}(x) - \nabla\tilde{\Lcal}(z)\|_2 & \leqslant \|\nabla \Lcal(x;y) - \nabla \Lcal(z;y)\|_2 + \|\nabla \Rcal_{\theta_1}(x) - \nabla \Rcal_{\theta_1}(z)\|_2 \\
        & \leqslant \|A^TA(x - z)\|_2 + L_1\|x-z\|_2 \\
        & \leqslant (\|A\|_2^2 + L_1)\|x-z\|_2.
        \end{align*}
        \item \textbf{Bound on majorization:} We claim that for each $t \in [T]$, if we set $\tilde{x}_t := x_t - \tilde{L}^{-1}\nabla q(x_t;x_t)$, then we have that bound $$q(x_t ; x_t) - q(\tilde{x}_t;x_t) \geqslant \frac{1}{2\tilde{L}}\|\nabla F(x_t)\|_2^2.$$ First, we note that since $\tilde{\Lcal}$ is $\tilde{L}$-smooth, we have that for any $x \in \R^d$, \begin{align}
            \tilde{\Lcal}(x) \leqslant \tilde{\Lcal}(x_t) + \langle \nabla \tilde{\Lcal}(x_t), x - x_t \rangle + \frac{\tilde{L}}{2}\|x - x_t\|_2^2. \label{eq:smoothness-Lcaltil}
        \end{align} By definition of $q(\cdot;x_t)$, we have that $\nabla q(x_t;x_t) = \nabla F(x_t)$ with function values \begin{align*}
            q(x_t;x_t) & = \tilde{\Lcal}(x_t)-\Rcal_{\theta_2}(x_t)\ \text{and} \\
            q(\tilde{x}_t;x_t) & = \tilde{\Lcal}(\tilde{x}_t) - \Rcal_{\theta_2}(x_t) + \tilde{L}^{-1}\langle g_t,\nabla F(x_t)\rangle.
        \end{align*} Using these properties of $q(\cdot;x_t)$ and applying \eqref{eq:smoothness-Lcaltil} with $x = \tilde{x}_t = x_t - \tilde{L}^{-1}\nabla q(x_t;x_t)= x_t - \tilde{L}^{-1}\nabla F(x_t)$, we see that \begin{align*}
            q(x_t;x_t) - q(\tilde{x}_t;x_t) & = \tilde{\Lcal}(x_t) - \tilde{\Lcal}(\tilde{x}_t) - \tilde{L}^{-1}\langle g_t, \nabla F(x_t)\rangle \\
            & \geqslant \langle \nabla \tilde{\Lcal}(x_t),\tilde{x}_t - x_t\rangle - \tilde{L}^{-1}\langle g_t, \nabla F(x_t)\rangle - \frac{\tilde{L}}{2}\|\tilde{x}_t - x_t\|_2^2 \\
            & = \tilde{L}^{-1} \langle \nabla \tilde{\Lcal}(x_t) - g_t, \nabla F(x_t)\rangle - \frac{1}{2\tilde{L}}\|\nabla F(x_t)\|_2^2 \\
            & = \left(\frac{1}{\tilde{L}} - \frac{1}{2\tilde{L}}\right)\|\nabla F(x_t)\|_2^2 = \frac{1}{2\tilde{L}}\|\nabla F(x_t)\|_2^2.
        \end{align*}
    \end{enumerate}
    Now that we have such bounds, we proceed with the proof of Theorem \ref{thm:dca-result}. Recall that $x_{t+1} \in \argmin_{x \in \R^d} q(x;x_t)$. Thus, $q(\tilde{x}_t ; x_t) \geqslant q(x_{t+1};x_t)$ for each $t \in [T]$. Moreover, we know that by the majorization properties of $q(\cdot;x_t)$, we have $F(x_t) = q(x_t;x_t)$ and $F(x) \leqslant q(x;x_t)$ for all $x \in \R^d$. Using such bounds, we get \begin{align*}
        F(x_t) - F(x_{t+1}) & \geqslant q(x_t;x_t) - q(x_{t+1};x_t) \geqslant q(x_t;x_t) - q(\tilde{x}_t;x_t) \geqslant \frac{1}{2\tilde{L}}\|\nabla F(x_t)\|_2^2
    \end{align*} where in the final inequality we used the bound on $q(\cdot;x_t)$ shown in the previous half of the proof. Summing both sides and dividing by $T+1$, we see that \begin{align*}
        \frac{1}{T+1}\sum_{t=0}^T\|\nabla F(x_t)\|_2^2 \leqslant \frac{2\tilde{L}(F(x_0) - F(x_{T+1}))}{T+1} \leqslant \frac{2\tilde{L}(F(x_0) -F_*)}{T+1}.
    \end{align*} Note that the above argument also shows that the objective values are decreasing at each iteration $F(x_{t+1}) \leqslant F(x_t) - (2\tilde{L})^{-1}\|\nabla F(x_t)\|_2^2$ and since we assumed $F$ is bounded below, the function values are convergent.

    To show stationarity of the limit point, the argument of \cite{khamaru2018convergence} shows that under our assumptions, continuity of both $\tilde{\Lcal}$ and its gradient $\nabla \tilde{\Lcal}$ guarantee that the subgradients associated to $(x_t)$ converge using graph continuity of the subdifferential \cite{rockafellar2009variational}.
\end{proof}

\begin{proof}[Proof of Theorem \ref{thm:prox-result}]
We first establish the following useful auxillary results for the proof:
\begin{itemize}
    \item \textbf{Subgradient representation:} We claim that $x_{t+1}$ has the following representation for some $b_{t+1} \in \partial \Rcal_{\theta_1}(x_{t+1})$: $$x_{t+1} = x_t - \alpha(\nabla \Lcal(x_t) + b_{t+1} - g_t).$$ This result will follow by noting that our update is the minimizer of a particular majorization of our objective. Define the function $h(\cdot;x_t)$ by \begin{align*}
        h(x;x_t) := \Lcal(x_t) - \Rcal_{\theta_2}(x_t) + \langle \nabla \Lcal(x_t) - g_t, x - x_t\rangle + \frac{1}{2\alpha}\|x-x_t\|_2^2 + \Rcal_{\theta_1}(x)
    \end{align*} where $g_t \in \partial \Rcal_{\theta_2}(x_t)$. Specifically, \begin{align*}
        x_{t+1} \in \mathrm{prox}^{\Rcal_{\theta_1}}_{1/\alpha}\left(x_t - \alpha(\nabla \Lcal(x_t) - g_t)\right) = \argmin_{x \in \R^d} h(x;x_t).
    \end{align*} Optimality guarantees that there exists a subgradient $b_{t+1} \in \partial \Rcal_{\theta_1}(x_{t+1})$ such that $$\nabla \Lcal(x_t) - g_t + b_{t+1} + \alpha^{-1}(x_{t+1} - x_t) = 0.$$
    \item \textbf{Objective decrease:} We claim that for each $t \in [T]$, $$F(x_t) - F(x_{t+1}) \geqslant \frac{1}{2\alpha}\|x_t-x_{t+1}\|_2^2.$$ Using convexity of $\Rcal_{\theta_1}$, we see that \begin{align*}
        F(x_t) - h(x_{t+1};x_t) & = \Lcal(x_t) + \Rcal_{\theta_1}(x_t) - \Rcal_{\theta_2}(x_t) - h(x_{t+1};x_t) \\
        & \geqslant \Lcal(x_t) - \Rcal_{\theta_2}(x_t) + \Rcal_{\theta_1}(x_{t+1}) + \langle b_{t+1},x_t - x_{t+1}\rangle - h(x_{t+1}; x_t) \\
        & = \langle \Lcal(x_t) - g_t + b_{t+1}, x_{t}-x_{t+1}\rangle - \frac{1}{2\alpha}\|x_t - x_{t+1}\|_2^2 \\
        & = \left(\frac{1}{\alpha} - \frac{1}{2\alpha}\right)\|x_t - x_{t+1}\|_2^2 = \frac{1}{2\alpha}\|x_t - x_{t+1}\|_2^2
    \end{align*} where we used the subgradient representation in the second to last equality. Using the fact that $h(\cdot;x_t)$ is a majorization of $F$, we conclude \begin{align}
        F(x_t) - F(x_{t+1}) \geqslant F(x_t) - h(x_{t+1};x_t) \geqslant \frac{1}{2\alpha}\|x_t - x_{t+1}\|_2^2. \label{eq:objective-decrease}
        \end{align}
    
\end{itemize}

Using the above results, we have the desired bound: \begin{align*}
        \frac{1}{T+1}\sum_{t=0}^T\|x_t - x_{t+1}\|_2^2 & \leqslant \frac{2\alpha}{T+1}\sum_{t=0}^T(F(x_t) - F(x_{t+1})) \\
        & = \frac{2\alpha(F(x_0)-F(x_{T+1}))}{T+1} \\
        & \leqslant \frac{2\alpha}{T+1}(F(x_0)-F_*). 
    \end{align*}
For the subsequent bound assuming smoothness of $\Rcal_{\theta_2}$, we have that by using the subgradient decomposition that \begin{align}
    \|\nabla F(x_{t+1})\|_2 & = \|\nabla \Lcal(x_{t+1}) - \nabla \Rcal_{\theta_2}(x_{t+1}) + b_{t+1}\|_2 \nonumber \\
    & =  \|\nabla \Lcal(x_{t+1}) - \nabla \Rcal_{\theta_2}(x_{t+1}) + (\nabla \Rcal_{\theta_2}(x_t) - \nabla \Lcal(x_t) + \alpha^{-1}(x_t - x_{t+1})) \|_2 \nonumber \\
    & \leqslant \|\nabla \Lcal(x_{t+1}) - \nabla \Lcal(x_t)\|_2 + \|\nabla \Rcal_{\theta_2}(x_{t+1}) - \nabla \Rcal_{\theta_2}(x_t)\|_2 + \alpha^{-1}\|x_{t+1} - x_t\|_2 \nonumber \\
    & \leqslant (\|A\|_2^2 + L_2 + \alpha^{-1})\|x_{t+1} - x_t\|_2. \label{eq:smoothness-bound}
\end{align} Combining this with the previous bound, we obtain \begin{align*}
    \frac{1}{T+1}\sum_{t=0}^{T}\|\nabla F(x_{t+1})\|_2^2 & \leqslant \frac{(\|A\|_2^2 + L_2 + \alpha^{-1})^2}{T+1}\sum_{t=0}^T\|x_{t+1}-x_t\|_2^2 \\
    & \leqslant \frac{2\alpha(\|A\|_2^2 + L_2 + \alpha^{-1})^2}{T+1}\cdot(F(x_0) - F_*).
\end{align*}

\paragraph{KL result:} For the final part of the proof, we first use Lemma 7 in \cite{khamaru2018convergence}, which states that under our assumptions, there exists $\omega \in [0,1)$, $\overline{F} \geqslant F_* > -\infty$, $M > 0$, and an integer $t_1$ such that for all $t \geqslant t_1$, we have \begin{align}
    |F(x_t) - \overline{F}|^{\omega} \leqslant M\|\nabla F(x_t)\|_2 \label{eq:kl-result}
\end{align} where $F(x_t) \downarrow \overline{F}.$ Using such a bound, we will first show that for $t \geqslant t_1$, \begin{align*}
    (F(x_{t}) - \overline{F})^{1 -  \omega} - (F(x_{t+1}) - \overline{F})^{1 - \omega}\geqslant \frac{(1- \omega)}{2M\alpha(\|A\|_2^2 + L_2 + \alpha^{-1})}\cdot \frac{\|x_t - x_{t+1}\|_2^2}{\|x_t - x_{t-1}\|_2}.
\end{align*} By concavity of the map $s \mapsto s^{1-\omega}$ and $F(x_t) \downarrow \overline{F}$, we have that for $t \geqslant t_1$,
\begin{align*}
    (F(x_{t}) - \overline{F})^{1 - \omega} - (F(x_{t+1}) - \overline{F})^{1 - \omega} & \geqslant (1- \omega)(F(x_t) - \overline{F})^{- \omega}(F(x_t) - F(x_{t+1})) \\
    & \geqslant (1- \omega)|F(x_t) - \overline{F}|^{- \omega} \cdot \frac{1}{2\alpha}\|x_t - x_{t+1}\|_2^2 \\
    & \geqslant \frac{1- \omega}{M\|\nabla F(x_t)\|_2}\cdot \frac{1}{2\alpha}\|x_t - x_{t+1}\|_2^2
\end{align*} where in the second inequality we used \eqref{eq:objective-decrease} and the last inequality we used \eqref{eq:kl-result}. Then we can use \eqref{eq:smoothness-bound} to lower bound $1/\|\nabla F(x_t)\|_2$ to obtain \begin{align*}
    (F(x_{t}) - \overline{F})^{1 - \omega} - (F(x_{t+1}) - \overline{F})^{1 - \omega}\geqslant \frac{1- \omega}{M(\|A\|_2^2 + L_2 + \alpha^{-1})}\cdot \frac{1}{2\alpha}\frac{\|x_t - x_{t+1}\|_2^2}{\|x_t - x_{t-1}\|_2}.
\end{align*} Now, for notational convenience, we set $\tilde{M}_{\alpha,\omega} := \frac{2M\alpha(\|A\|_2^2 + L_2 + \alpha^{-1})}{1-\omega}.$ Note that this inequality can be rewritten as \begin{align*}
    \|x_{t} - x_{t+1}\|_2^2 \leqslant \tilde{M}_{\alpha,\omega}\cdot \left[(F(x_{t}) - \overline{F})^{1 - \omega} - (F(x_{t+1}) - \overline{F})^{1 - \omega})\right]\|x_t - x_{t-1}\|_2,\ \forall t \geqslant t_1.
\end{align*} In \cite{khamaru2018convergence}, the authors show using a weighted AM-GM inequality that this bound gives for $t \geqslant t_1 + 2$, \begin{align*}
    2 \sum_{\ell = t_1 + 1}^t \|x_{\ell} - x_{\ell+1}\|_2 \leqslant \tilde{M}_{\alpha,\omega}(F(x_{t_1+1}) - \overline{F})^{1-\omega} + \sum_{\ell=t_1 + 1}^t \|x_{\ell} - x_{\ell-1}\|_2.
\end{align*} One can then further rewrite this bound for $t \geqslant t_1 + 2$ as \begin{align*}
    \sum_{\ell=t_1+1}^t\|x_{\ell} - x_{\ell+1}\|_2 \leqslant \tilde{M}_{\alpha,\omega}(F(x_{t_1+1}) - \overline{F})^{1-\omega} + \|x_{t_1} - x_{t_1+1}\|_2.
\end{align*} Taking the limit on the left-hand side yields the following: \begin{align}
     \sum_{t=t_1+1}^{\infty}\|x_t - x_{t+1}\|_2 \leqslant \tilde{M}_{\alpha,\omega}(F(x_{t_1+1}) - \overline{F})^{1-\omega} + \|x_{t_1} - x_{t_1+1}\|_2 \label{eq:final-tail-bound}
\end{align} Now, we can exploit \eqref{eq:smoothness-bound} and \eqref{eq:final-tail-bound}: letting $\tilde{M} := \|A\|_2^2 + L_2+\alpha^{-1}$, we see that for $T \geqslant t_1$, \begin{align*}
    \sum_{t=0}^{T} \|\nabla F(x_t)\|_2 & \leqslant \tilde{M}\sum_{t=0}^T\|x_t - x_{t-1}\|_2 \\
    & \leqslant \tilde{M}\cdot\left(\sum_{t=0}^{t_1+1}\|x_t - x_{t-1}\|_2 + \sum_{t=t_1+1)}^{\infty}\|x_t - x_{t-1}\|_2\right) \\
    & \leqslant \tilde{M}\cdot\left(\sum_{t=0}^{t_1+1}\|x_t - x_{t-1}\|_2 +  \tilde{M}_{\alpha,\omega}(F(x_{t_1+1}) - \overline{F})^{1-\omega} + \|x_{t_1} - x_{t_1+1}\|_2\right) =: c_{\omega}
\end{align*} where we used \eqref{eq:smoothness-bound} in the first inequality and \eqref{eq:final-tail-bound} in the last inequality. Dividing both sides by $T+1$ yields the linear rate \begin{align*}
    \frac{1}{T+1}\sum_{t=0}^T\|\nabla F(x_t)\|_2 \leqslant \frac{c_{\omega}}{T+1}.
\end{align*}
\end{proof}

\section{Implementation Details}

Experiments were conducted on a Linux-based system with  CUDA 12.2 equipped with 4 Nvidia 
A6000 GPUs, each of them has 48GB of memory. 
\subsection{Distance function approximation}
To handle 2-dim data, we replace the convolutional layers used in the CT experiments with linear layers. Specifically, the hidden dimension of linear layers is 200. All the other implementation settings are the same as those in the CT experiments. 
%%%%%%%%%%%%%%%%%%%%%%%%%%%
\subsection{CT experiments}
All of the methods are implemented in PyTorch based on the official repository of CT\footnote{\url{https://github.com/Zakobian/CT_framework_/tree/master}}. 
The LPD method is trained on pairs of target images and projected data, while the U-Net post-processor is trained on pairs of true images and the corresponding reconstructed FBPs. The TV method was computed by employing the ADMM-based solver in ODL while the FBP was also computed with the help of ODL.

\begin{figure}[h]
    \centering
 
    \includegraphics[width=0.75\linewidth]{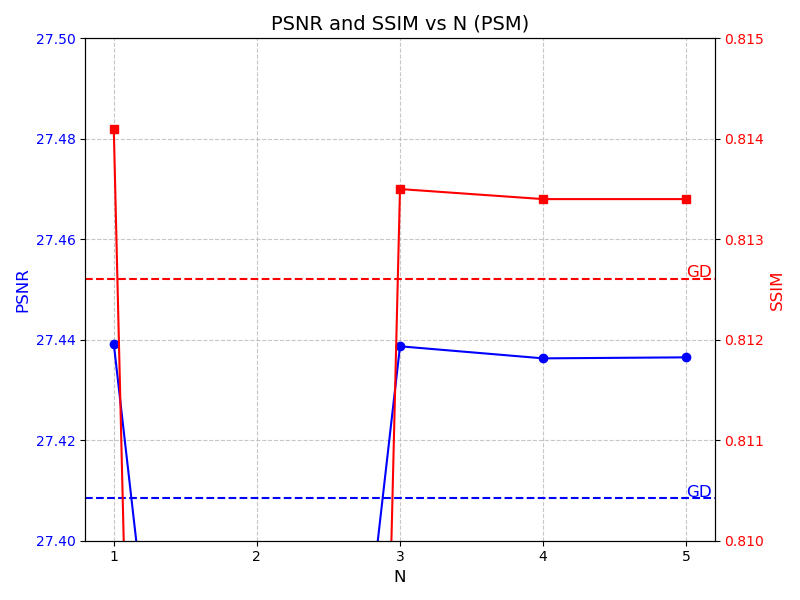}
    \caption{\textbf{Ablation on inner-loop iteration number $N$ of PSM  in the limited-view setting.}   PSM consistently improves the ADCR across various choices of the inner-loop iteration number $N$. Results in the sparse-view setting are similar. The dashed lines represent the results of ADCR obtained through gradient descent.    For PSM, we choose $N=1$ for both settings. }
    \label{fig:abl-psm}
\end{figure}

\begin{figure}[!h]
    \centering
    \includegraphics[width=0.75\linewidth]{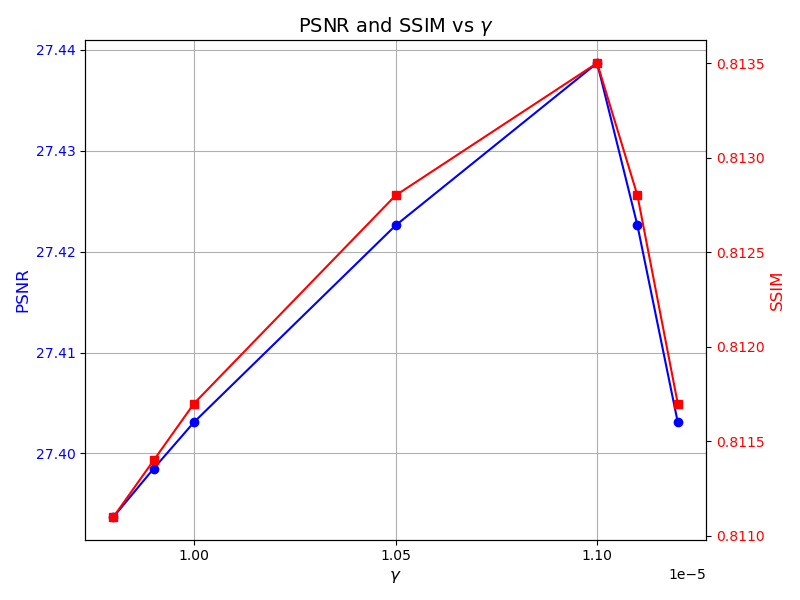}
    \caption{\textbf{Ablation on the proximal strength $\gamma$ of PSM algorithm in the limited-view setting.} Results in the sparse-view setting are similar.   We choose $\gamma=1.1e-5$ in our implementation. }
    \label{fig:abl-lambda}
\end{figure}

\begin{figure*}[h]
    \centering
    \includegraphics[width=1\linewidth]{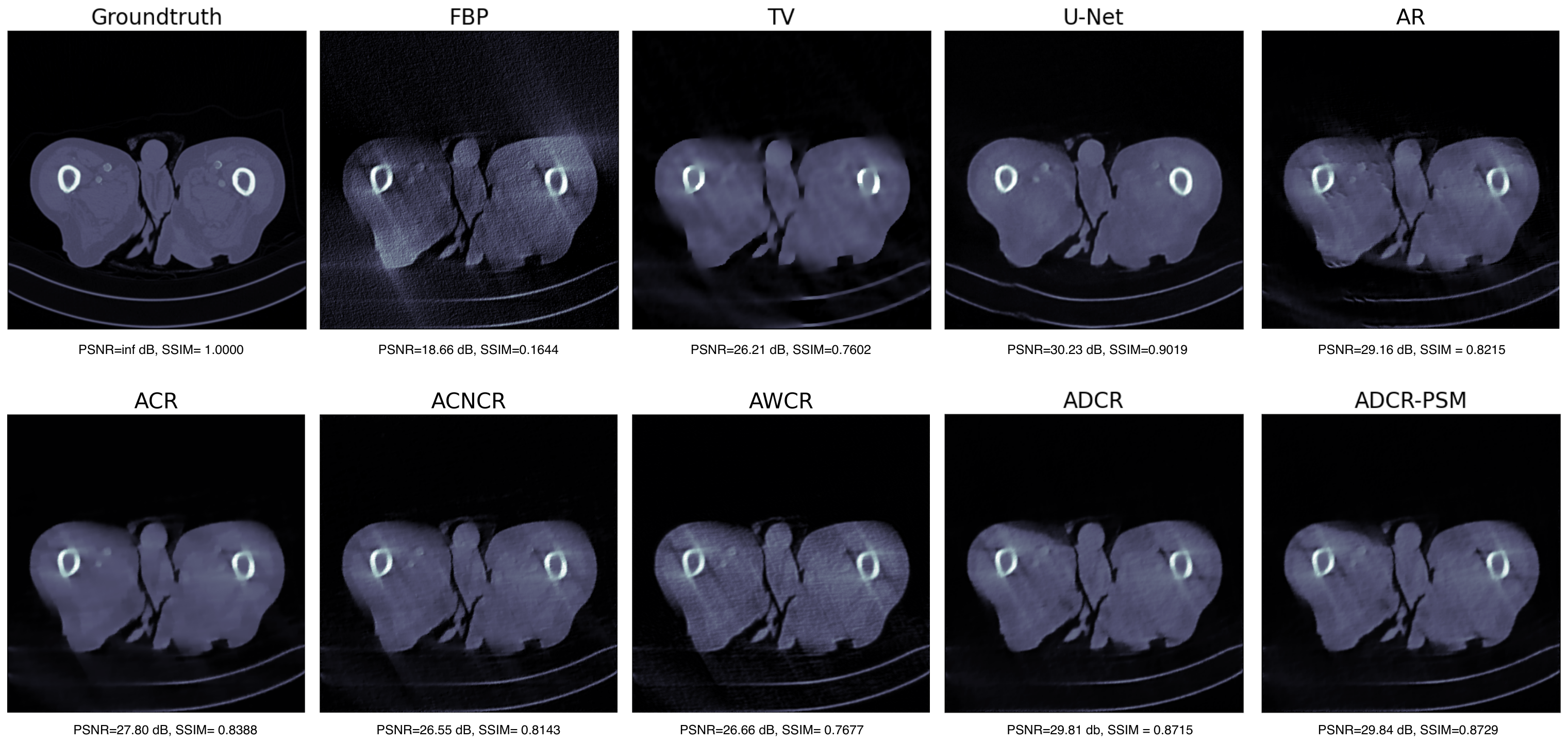}
    \caption{Reconstructed images obtained using different methods, along with the associated PSNR and SSIM, for limited view CT.}
    \label{fig:limited}
\end{figure*}

\paragraph{More ablations  in PSM} 
\begin{enumerate}
\item \textbf{Inner-loop iteration number $N$.} As shown in Figure \ref{fig:abl-psm}, a noticeable improvement in both PSNR and SSIM occurs for all values of $N$ in PSM, except for $N=2$. By default, we choose the optimal $N=1$ for both the sparse-view and limited-view settings.
    \item \textbf{Proximal strength $\gamma$. } We explore the impact of varying the hyperparameter $\gamma$ within the proposed PSM framework. By systematically adjusting $\gamma$, we assess its effect on the PSNR and SSIM of our method. The results, illustrated in Figure \ref{fig:abl-lambda}, shows that the tuning of $\gamma$ is effective and by default, we choose $\gamma=1.1e-5$ in our experiments. 
\end{enumerate}

\paragraph{Qualitative comparison in the limited-view setting:} We provide qualitative comparison of the reconstructed images obtained from various methods in the limited-view setting. As illustrated in Figure \ref{fig:limited}, ADCR yields   superior results in terms of both PSNR and SSIM to other weakly supervised methods. 

\paragraph{\#Params of NNs:} The number of parameters of neural networks used in different baselines are documented in Table \ref{tab:params}, which mostly aligns with the Table 1 in \cite{shumaylov2023provably} except that 
  we enlarge the neural network used in AR in the limited-view setting and ACR/AWCR in the sparse-view setting to achieve better performance.

\subsubsection{Architecture design}

\paragraph{AR \cite{lunz2018adversarial}} Following the implementation of the original paper, the neural network used in AR is a standard convolutional neural network.  It consists of six convolutional layers with $5 \times 5$ kernels, padding of 2, and increasing channel sizes from 16 to 128, doubled at each layer. Strided convolutions are applied from the third layer onward, progressively downsampling the spatial resolution.  Each layer is followed by a LeakyReLU activation with a negative slope of 0.1. The downsampled feature maps are flattened and passed to a fully connected layer with 256 units, followed by another linear layer that produces a scalar output.

%It has six convolutional layers (five intermediate layers and one final layer) designed for processing single-channel 2D inputs. Each convolutional layer employs a kernel size of 5 with a stride of 1 and padding of 2, preserving spatial dimensions throughout the network. The input layer maps the single-channel input to 32 feature channels. The intermediate layers alternate between two types of convolutional operations: one set processes the input features directly, while the other transforms the intermediate feature maps, both followed by a LeakyReLU activation with a negative slope of 0.2. The final convolutional layer reduces the output to a single channel, and the spatial dimensions are averaged globally to produce a scalar output for each input sample.

\paragraph{ACR \cite{acr}} Following the implementation of the original paper, the network is an ICNN tailored for 2D inputs. Each convolutional layer has a kernel size of 5, stride 1, and padding of 2, ensuring preserved spatial dimensions. The network uses LeakyReLU activations with a negative slope of 0.2 after each layer.  It supports two configurations: in the  {sparse-view setting}, the network has 48 feature channels and 10 intermediate layers, providing enhanced capacity to process sparse input patterns, with a regularization parameter $\mu = 5$. In the  {limited-view setting}, the network employs 16 feature channels and 5 intermediate layers, to avoid overfitting, and uses $\mu = 10$. Both settings globally average the final feature maps to produce a scalar output per input, with weight clamping applied to enforce convexity. 

\paragraph{ACNCR \cite{shumaylov2023provably}} 
Following the implementation of the original paper, the usual ICNN architecture with LeakyReLU activations is used. The smooth network is parametrized as a deep convolutional network with SiLU activations and five layers. The RMSprop optimizer with a learning rate of $1 \times 10^{-4}$ is employed for training.

\paragraph{AWCR \cite{shumaylov2024weakly}}  We follow the setting of the original paper except that the starting channel size of the ICNN component in the sparse-view setting is increased to 128. For sparse-view CT, the ICNN component of the AWCR comprises five convolutional layers with LeakyReLU activations, $5 \times 5$ kernels, and 128 channels. The smooth component includes six convolutional layers with SiLU activations, $5 \times 5$ kernels, and a doubling channel configuration starting at 16, using a stride of 2, and ending with 128 channels in the final layer. For limited-view CT, the ICNN component remains the same except that the starting channel size is 16 while the smooth component is simplified to a single convolutional layer with a $7 \times 7$ kernel, SiLU activation, and 32 channels, based on insights that reducing depth and channels minimizes overfitting in limited-angle settings.  

\paragraph{ADCR (ours)} The two ICNNs parametrized by $\theta_1$ and $\theta_2$ in our algorithm are identical. For the sparse-view setting, the ICNN component of the ADCR is constructed using four convolutional layers with LeakyReLU activations and 5$\times$5 kernels. The channel size doubles at each layer, starting from 16. The LeakyReLU's slope is 0.2. For the limited-view setting, the ICNN component of the ADCR is constructed using 10 convolutional layers with LeakyReLU activations and 5$\times$5 kernels. The channel size remains the same as 32 at all layers. The LeakyReLU's slope is 0.2.
We use the Adam optimizer with a learning rate of $5 \times 10^{-5}$.

\begin{table}[h]
\caption{\textbf{The number of parameters of neural networks used in different methods.} Compared with Table 1 in \cite{shumaylov2023provably}, we enlarge the neural network used in AR in the limited-view setting and ACR/AWCR in the sparse-view setting to achieve better performance. } \label{tab:params}
    \centering
    \begin{tabular}{l l l}
    \toprule
        ~ & \multicolumn{2}{c}{\# Params}  \\  
        \cmidrule{2-3}
        Methods & Limited & Sparse \\ 
        \midrule
        LPD & 127,370 & 700,180\\ \midrule
        U-Net & 14,787,777 & 14,787,777 \\ \midrule
        AR & 33,952,481 & 33,952,481 \\ \midrule
        ACR & 34,897 & 590,929 \\ \midrule
        ACNCR & 1,085,448 & 1,085,448 \\ \midrule
        AWCR & 343,025 &  4,907,105 \\  \midrule
%        AD(W)CR & 686,049 & 9,814,209\\ \midrule
        ADCR & 531,904 & 8,781,184 \\ 
        \bottomrule
    \end{tabular}
\end{table}

\end{document}